\documentclass{article}

\usepackage[final,nonatbib]{neurips_2022}

\usepackage[utf8]{inputenc} 
\usepackage[T1]{fontenc}    
\usepackage{hyperref}       
\usepackage{url}            
\usepackage{booktabs}       
\usepackage{amsfonts}       
\usepackage{nicefrac}       
\usepackage{microtype}      
\usepackage{xcolor}         
\usepackage{marvosym}

\usepackage[numbers]{natbib}

\usepackage{graphicx}
\usepackage{subfigure}
\usepackage{multirow}
\usepackage{amsmath}
\usepackage{amsfonts,amssymb}

\newcommand{\R}{\mathbb R}

\DeclareMathOperator*{\argmin}{arg\,min}

\newtheorem{scenario}{Scenario}
\newtheorem{remark}{Remark}

\title{Meta-Auto-Decoder for Solving Parametric Partial Differential Equations}

\author{%
Xiang Huang\thanks{The first two authors contributed equally to this paper, and Bin Dong is the
corresponding author.
Zhanhong Ye proposed MAD-L and explain the effectiveness of the MAD method from the perspective of manifold learning. 
Huang Xiang proposed MAD-LM on the basis of MAD-L and completed all the experiments in the paper. Xiang Huang performed this work during an internship at
Huawei.} \\
\texttt{sahx@mail.ustc.edu.cn} \\
University of Science and \\ 
Technology of China \\
\And
Zhanhong Ye$^*$ \\
\texttt{yezhanhong@pku.edu.cn} \\
Peking University \\
\And
Hongsheng Liu \\
\texttt{liuhongsheng4@huawei.com} \\
Huawei Technologies Co. Ltd \\
\And
Beiji Shi \\
\texttt{shibeiji@huawei.com} \\
Huawei Technologies Co. Ltd \\
\And
Zidong Wang \\
\texttt{wang1@huawei.com} \\
Huawei Technologies Co. Ltd \\
\And
Kang Yang \\
\texttt{yangkang22@huawei.com} \\
Huawei Technologies Co. Ltd \\
\And
Yang Li \\
\texttt{liyang477@huawei.com} \\
Huawei Technologies Co. Ltd \\
\And
Min Wang \\
\texttt{wangmin106@huawei.com} \\
Huawei Technologies Co. Ltd \\
\And
Haotian Chu \\
\texttt{chuhaotian2@huawei.com} \\
Huawei Technologies Co. Ltd \\
\And
Fan Yu \\
\texttt{fan.yu@huawei.com} \\
Huawei Technologies Co. Ltd \\
\And
Bei Hua \\
\texttt{bhua@ustc.edu.cn} \\
University of Science and \\
Technology of China \\
\And
Lei Chen \\
\texttt{leichen@cse.ust.hk} \\
Hong Kong University of \\ 
Science and Technology \\
\And
Bin Dong\textsuperscript{\Letter} \\
\texttt{dongbin@math.pku.edu.cn}\\
Beijing International Center for Mathematical Research, Peking University \\
Center for Machine Learning Research, Peking University \\
}

\begin{document}

\maketitle

\begin{abstract}
Many important problems in science and engineering require solving the so-called parametric partial differential equations (PDEs), i.e., PDEs with different physical parameters, boundary conditions, shapes of computation domains, etc.  Recently, building learning-based numerical solvers for parametric PDEs has become an emerging new field.  One category of methods such as the Deep Galerkin Method (DGM) and Physics-Informed Neural Networks (PINNs) aim to approximate the solution of the PDEs. They are typically unsupervised and mesh-free, but require going through the time-consuming network training process from scratch for each set of parameters of the PDE.  Another category of methods such as Fourier Neural Operator (FNO) and Deep Operator Network (DeepONet) try to approximate the solution mapping directly.  Being fast with only one forward inference for each PDE parameter without retraining, they often require a large corpus of paired input-output observations drawn from numerical simulations, and most of them need a predefined mesh as well.  In this paper, we propose Meta-Auto-Decoder (MAD), a mesh-free and unsupervised deep learning method that enables the pre-trained model to be quickly adapted to equation instances by implicitly encoding (possibly heterogenous) PDE parameters as latent vectors.  The proposed method MAD can be interpreted by manifold learning in infinite-dimensional spaces, granting it a geometric insight.  Extensive numerical experiments show that the MAD method exhibits faster convergence speed without losing accuracy than other deep learning-based methods.
The project page with code is available: \href{https://gitee.com/mindspore/mindscience/tree/master/MindElec/}{https://gitee.com/mindspore/mindscience/tree/master/MindElec/}.
\end{abstract}

\section{Introduction}\label{sec:intro}

\noindent
Many important problems in science and engineering, such as inverse problems, control and optimization, risk assessment, and uncertainty quantification~\cite{cohen2015approximation,khoo2021solving}, require solving the so-called parametric PDEs, i.e., partial differential equations (PDEs) with different physical parameters, boundary conditions, or solution regions.
Mathematically, they require to solve the so-called \textit{parametric} PDEs that can be formulated as:
\begin{equation}\label{def:PDE}
	\mathcal{L}_{\widetilde{x}}^{\gamma_1} u = 0, \ {\widetilde{x}} \in \Omega \subset \R^d,\qquad
	\mathcal{B}_{\widetilde{x}}^{\gamma_2} u = 0, \ {\widetilde{x}} \in \partial \Omega
\end{equation}
where $\mathcal{L}^{\gamma_1}$ and $\mathcal{B}^{\gamma_2}$ are partial differential operators parametrized by $\gamma_1$ and $\gamma_2$, respectively, and ${\widetilde{x}}$ denotes the independent variable in spatiotemporal-dependent PDEs. 
Given $\mathcal{U} = \mathcal{U}(\Omega;\R^{d_u})$ and the space of parameters $\mathcal{A}$, $\eta=(\gamma_1,\gamma_2,\Omega) \in \mathcal{A}$ is the variable parameter of the PDEs and $u \in \mathcal{U}$ is the solution of the PDEs. 
Note that the form of $\eta$ considered here is very general with possible heterogeneity allowed, since the computational domain shape $\Omega$ and the functions defined on this domain or its boundary (which may be involved in $\gamma_1,\gamma_2$) is obviously of different type. 
Solving parametric PDEs requires to learn an infinite-dimensional operator $G:\mathcal{A} \to \mathcal{U}$ that map any PDE parameter $\eta$ to its corresponding solution $u^\eta$ (i.e., the solution mapping).


In recent years, learning-based PDE solvers have become very popular, and it is generally believed that learning-based PDE solvers have the potential to improve efficiency~\cite{raissi2018deep,ChiyuMaxJiang2020MeshfreeFlowNetAP,DmitriiKochkov2021MachineLA}.
The learning-based PDE solvers can be categorized into two categories in terms of the objects that are approximated by neural networks (NN), i.e., the approximation of the solution $u^\eta$ and the approximation of the solution mapping $G$.

\paragraph{NN as a new ansatz of solution.}
This kind of approaches approximate the solution of the PDEs with a neural network and mainly rely on governing equations and boundary conditions (or their variants) to train the neural networks. 
For example, PINNs~\cite{raissi2018deep} and DGM~\cite{sirignano2018dgm} constrain the output of deep neural networks to satisfy the given governing equations and boundary conditions. 
Deep Ritz Method (DRM)~\cite{weinan2018deep} exploits the variational form of PDEs and can be used to solve PDEs that can be reformulated as equivalent energy minimization problems. 
Based on a weak formulation of PDEs, Weak Adversarial Network (WAN)~\cite{zang2020weak} parameterizes the weak solution and test functions as primal and adversarial neural networks, respectively. 
These neural approximation methods can work in an unsupervised manner, without the need to generate labeled data from conventional computational methods. 
However, all these methods treat different PDE parameters as independent tasks, and need to retrain the neural network from scratch for each PDE parameter. 
When a large number of tasks with different PDE parameters need to be solved, these methods are computationally expensive and impractical.
In order to mitigate retraining cost, \citet{weinan2018deep} recommends a transfer learning method that uses a model trained for one task as the initial model to train another task. 
However, according to our experiments, transfer learning method is not always effective in improving convergence speed (see Sec.\ref{sec:burgers}, \ref{sec:laplace}). 

\paragraph{NN as a new ansatz of solution mapping.}
This kind of approaches use neural networks to learn the solution mapping between two infinite-dimensional function spaces~\cite{long2018pde,long2019pde,lu2019deeponet, bhattacharya2020model, li2020fourier}.
For example, PDE-Nets~\cite{long2018pde,long2019pde} are among the earliest neural operators that are specifically designed convolutional neural networks inspired by finite difference approximations of PDEs. 
They are able to uncover hidden PDE models from observed dynamical data and perform fast and accurate  predictions at the same time.
DeepONet~\cite{lu2019deeponet} uses two subnets to encode the parameters and location variables of the PDEs separately, and merge them together to compute the solution.
FNO~\cite{li2020fourier} utilizes fast Fourier transform to build the neural operator architecture and learn the solution mapping between two infinite-dimensional function spaces.
A significant advantage of these approaches is that once the neural network is trained, the prediction time is almost negligible. 
Although they have demonstrated promising results across a wide range of applications, several issues occur. 
First, the data acquisition cost is prohibitive in complex physical, biological, or engineering systems, and the generalization ability of these models is poor when there is not enough labeled data~\cite{cai2021physics}.
Second, most of these methods~\cite{long2018pde,long2019pde,bhattacharya2020model,li2020fourier} require a predefined mesh and utilize the labeled data on the mesh for training and inference. 
Third, simply applying one forward inference may lead to unsatisfactory generalization, especially on out-of-distribution (OOD) settings (i.e., PDE parameters for training and inference are from different probability distributions).
Finally, these operators directly takes the PDE parameter $\eta$ as network input, which would bring inconvenience in network implementation if $\eta$ is heterogeneous.
The recently proposed Physics-Informed DeepONet (PI-DeepONet)~\cite{wang2021learning} can learn a mesh-free solution mapping without any labeled data and retraining.
However, it needs to collect a large number of training samples in the parameter space $\mathcal{A}$ to obtain an acceptable accuracy (see Sec.\ref{sec:burgers}), and is still inflexible dealing with heterogeneous PDE parameters. 

\paragraph{Meta-Learning.}
Different from conventional machine learning that learns to do a given task, meta-learning learns to improve the learning algorithm itself based on multiple learning episodes over a distribution of related tasks.
As a result, meta-learning can handle new tasks faster and better. 
In this field, the Model-Agnostic Meta-Learning (MAML)~\cite{finn2017model} algorithm and its variants~\cite{antoniou2019train, nichol2018reptile, yoon2018bayesian} have beed widely used. 
These algorithms try to find an initial model with good generalization ability such that it can be adapted to new tasks with a small number of gradient updates.  
For example, MAML~\cite{finn2017model} firstly trains a meta-model with good initialization weight on a variety of learning tasks, which is then fine-tuned on a new task through a few steps of gradient descent to get the target model.
The Reptile~\cite{nichol2018reptile} algorithm eliminates second-order derivatives in MAML algorithm by repeatedly sampling a task, training on it, and moving the initialization towards the trained weight on that task.

Borrowing the idea of meta-learning may inspire new ways to solve parametric PDEs, where different PDE parameters correspond to different tasks. 
To the best of our knowledge, Meta-MgNet~\cite{chen2022meta} is the first work that view solving parametric PDEs as a meta-learning problem, which is based on hypernet and the multigrid algorithm. 
Meta-MgNet utilizes the similarity between tasks to generate good smoothing operators adaptively, and thereby accelerates the solution process, but is not directly applicable to PDEs on which the multigrid algorithm is not available. 
Recently, the Reptile algorithm is also used to accelerate the PDE solving problems in~\cite{liu2021novel}.
However, MAML and Reptile are not always effective in improving convergence speed (see Sec.\ref{sec:maxwell} and \ref{sec:laplace}). 

\paragraph{Our contributions.}
We propose Meta-Auto-Decoder (MAD), a mesh-free and unsupervised deep learning method that enables the pre-trained model to be quickly adapted to equation instances by implicitly encoding heterogeneous PDE parameters as latent vectors.
Different from Meta-MgNet, MAD makes use of the similarity between tasks from the perspective of manifold learning, and tries to learn a nonlinear approximation of the solution manifold. 
We construct the ansatz of solution as a neural network in the form $u_\theta({\widetilde{x}},z)$. 
By taking the spatial (or spatial-temporal) coordinate ${\widetilde{x}}$ directly as the network input, unsupervised training loss is allowed, and a mesh is no longer required. 
As the additional input $z$ varies, $u_\theta({\widetilde{x}},z)$ moves on a manifold in an infinite-dimensional function space, which may be an approximation of the true solution manifold for certain $\theta$. 
The PDE parameter $\eta$ is implicitly encoded into $z$ by applying the auto-decoder architecture motivated by \cite{park2019deepsdf}, regardless of the possible heterogeneity. 
When a new task comes, MAD achieves fast transfer by projecting the new task to the manifold and fine-tuning the manifold at the same time.
The main contributions of this paper are summarized as follows: 
\begin{itemize}
\item A mesh-free and unsupervised deep neural network approach is proposed to solve parametric PDEs. Based on meta-learning concept, once the neural network is pre-trained, solving a new task involves only a small number of iterations. In addition, the auto-decoder architecture adopted by MAD can realize auto-encoding of heterogeneous PDE parameters.
\item The mathematical intuition behind the MAD method is analyzed from the perspective of manifold learning. In short, a neural network is pre-trained to approximate the solution manifold, and the required solution is searched on the solution manifold or in a neighborhood of the solution manifold.
\item Extensive numerical experiments are carried out to demonstrate the effectiveness of our method, which show that MAD can significantly improve the convergence speed and has good extrapolation ability for OOD settings.
\end{itemize}

\section{Methodology}\label{sec:methodology}

\subsection{Meta-Auto-Decoder}

\begin{figure}
\begin{center}
\centerline{\includegraphics[width=0.8\columnwidth]{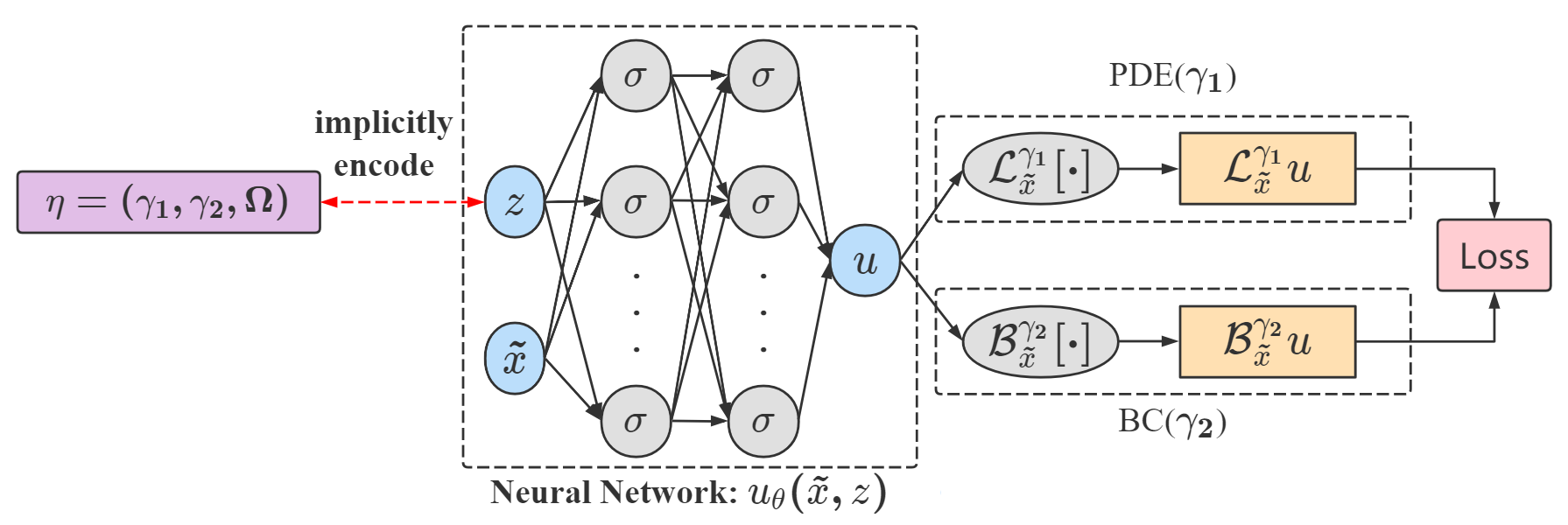}}
\caption{Architecture of Meta-Auto-Decoder.}
\label{fig:AD_DNN}
\end{center}
\end{figure}

We adopt meta-learning concept to realize fast solution of parametric PDEs. 
Our basic idea is to first learn some universal meta-knowledge from a set of sampled tasks in the pre-training stage, and then solve a new task quickly by combining the task-specific knowledge with the shared meta-knowledge in the fine-tuning stage. 
We also adapt the auto-decoder architecture in~\cite{park2019deepsdf}, 
and introduce $u_\theta({\widetilde{x}},z)$ to approximate the solutions of parametric PDEs. 
The architecture of $u_\theta({\widetilde{x}},z)$ is shown in Fig.\ref{fig:AD_DNN}.
A physics-informed loss is used for training, making the proposed method unsupervised.
Putting all these together, we propose a new method Meta-Auto-Decoder (MAD) to solve parametric PDEs. 
For the rest of the subsection, the loss function and the two stages of training will be explained in details. 

To enable unsupervised learning, 
given any PDE parameter $\eta\in\mathcal{A}$, the physics-informed loss $L^\eta:\mathcal{U}\to[0,\infty)$ about Eq.\eqref{def:PDE}
\begin{equation}
	L^\eta[u] = \|\mathcal{L}_{\widetilde{x}}^{\gamma_1} u\|_{L^2(\Omega)}^2 + \lambda_\text{bc}\|\mathcal{B}_{\widetilde{x}}^{\gamma_2} u\|_{L^2(\partial\Omega)}^2
\end{equation}
is considered, where $\lambda_\text{bc}>0$ is a weighting coefficient.
The Monte Carlo estimate of $L^\eta[u]$ is
\begin{equation}\label{eq:MCPIloss}
	\hat L^\eta[u]
	=\frac1{M_\text{r}}\sum_{j=1}^{M_\text{r}}\Bigl\|\mathcal{L}_{\widetilde{x}}^{\gamma_1} u({\widetilde{x}}_j^\text{r})\Bigr\|_2^2 +
	\frac{\lambda_\text{bc}}{M_\text{bc}}\sum_{j=1}^{M_\text{bc}}\Bigl\|\mathcal{B}_{\widetilde{x}}^{\gamma_2} u({\widetilde{x}}_j^\text{bc})\Bigr\|_2^2
,\end{equation}
where $\{{\widetilde{x}}_j^\text{r}\}_{j\in\{1,\dots,M_\text{r}\}}$ and $\{{\widetilde{x}}_j^\text{bc}\}_{j\in\{1,\dots,M_\text{bc}\}}$ are two sets of random sampling points in $\Omega$ and $\partial\Omega$, respectively. 
This task-specific loss $\hat L^\eta[u]$ can be computed by automatic differentiation~\cite{baydin2018automatic},
and will be used in the pre-training stage and the fine-tuing stage. 

In the pre-training stage, through minimizing the loss function, a pre-trained model parametrized by $\theta^*$ is learned for all tasks and each task is paired with its own decoded latent vector $z_i^*$.
Such a pre-trained model is considered as the meta knowledge as it is learned from the distribution of all tasks and the learned latent vector $z_i^*$ is the task-specific knowledge. 
When solving a new task in the fine-tuning stage, keep the model weight $\theta^*$ fixed and minimize the loss by fine-tuning the latent vector $z$.
Alternatively, we may unfreeze $\theta$ and allow it to be fine-tuned along with $z$. 
These two fine-tuning strategies give rise to different versions of MAD, which are called \textit{MAD-L} and \textit{MAD-LM}, respectively. 
The corresponding problems of pre-training and fine-tuning are formulated as follows:

\paragraph{Pre-training Stage}
Given $N$ randomly generated PDE parameters $\eta_1,\dots,\eta_N\in\mathcal{A}$, both \textit{MAD-L} and \textit{MAD-LM} solve the following optimization problem
\begin{equation}\label{eq:MADtr}
	(\{z^*_i\}_{i\in\{1,\dots,N\}},\; \theta^*) = \operatorname*{\arg\min}_{\theta,\{z_i\}_{i\in\{1,\dots,N\}}}\sum_{i=1}^{N}\left(\hat L^{\eta_i}[u_\theta(\cdot,z_i)]+\frac1{\sigma^2}\|z_i\|^2\right),
\end{equation}
where $\theta^*$ is the optimal model weight, $\{z_i^*\}_{i\in\{1,\dots,N\}}$ are the optimal latent vectors for different PDE parameters, and $\hat L^{\eta_i}$ is defined in Eq.\eqref{eq:MCPIloss}.
The regularization $\frac1{\sigma^2}\|z_i\|^2$ is added for training stability.

\paragraph{Fine-tuning Stage (\textit{MAD-L})}
Given a new PDE parameter $\eta_{\text{new}}$, 
\textit{MAD-L} keeps $\theta^*$ fixed, and minimizes the following loss function to get
\begin{equation}\label{eq:MADinf}
	z_{\text{new}}^*=\operatorname*{\arg\min}_{z}\hat L^{\eta_{\text{new}}}[u_{\theta^*}(\cdot,z)]+\frac1{\sigma^2}\|z\|^2
,\end{equation}
then $u_{\theta^*}(\cdot,z_{\text{new}}^*)$ is the approximate solution of PDEs with parameter $\eta_{\text{new}}$.
To speed up convergence, we can set the initial value of $z$ to $z_i^*$ obtained during pre-training where $\eta_i$ is the nearest%
\footnote{
For example, if $\mathcal{A}$ is a space of functions, we can discretize a function into a vector and then find the Euclidean distance between the two vectors as the distance between two PDE parameters.}
to $\eta_{\text{new}}$.

\paragraph{Fine-tuning Stage (\textit{MAD-LM})}
\textit{MAD-LM} fine-tunes the model weight $\theta$ with the latent vector $z$ simultaneously, and solves the following optimization problem
\begin{equation}\label{eq:MADinf_LM}
	(z_{\text{new}}^*,\theta_{\text{new}}^*)=\operatorname*{\arg\min}_{z,\theta}\hat L^{\eta_{\text{new}}}[u_{\theta}(\cdot,z)]+\frac1{\sigma^2}\|z\|^2
\end{equation}
with initial model weight $\theta^*$. 
This would produce an alternative approximate solution $u_{\theta_{\text{new}}^*}(\cdot,z_{\text{new}}^*)$. 
The latent vector is initialized in the same way as \textit{MAD-L}. 
\begin{remark}
	The MAD method has several key advantages compared with existing methods. 
	Besides being mesh-free and unsupervised, it can deal with heterogeneous PDE parameters painlessly, since $\eta$ is not taken as the network input, and is encoded into $z$ in an implicit way. 
	Introduction of the meta-knowledge $\theta^*$ would accelerate the fine-tuning process, which can be better understood in the light of the manifold learning perspective. 
	For \textit{MAD-LM}, the accuracy on OOD tasks is likely to be at least comparable with training from scratch based on PINNs. 
	Although the fine-tuning process of MAD is still slower than one forward inference of a neural network solution mapping, the advantages presented above can make it more suitable for some real applications. 
\end{remark}

\begin{remark}
	If we replace the physics-informed loss by certain supervised loss, the \textit{MAD-L} method would then coincide with the DeepSDF algorithm~\cite{park2019deepsdf}. 
	Despite of this, the field of solving parametric PDEs is quite different from 3D shape representation in computer graphics. 
	Moreover, the introduction of model weight fine-tuning in \textit{MAD-LM} can significantly improve solution accuracy, as is explained intuitively in Sec.\ref{sec:MAD_L},\ref{sec:MAD_LM} and validated by numerical experiments in Sec.\ref{sec:numerical_experiments}. 
\end{remark}

\subsection{Manifold Learning Interpretation of \textit{MAD-L}}\label{sec:MAD_L}
\begin{figure*}
\centering
\subfigure[Fine-tune $z$ only (\textit{MAD-L})]{
	\label{fig:MAD}
	\includegraphics[width=0.42\columnwidth]{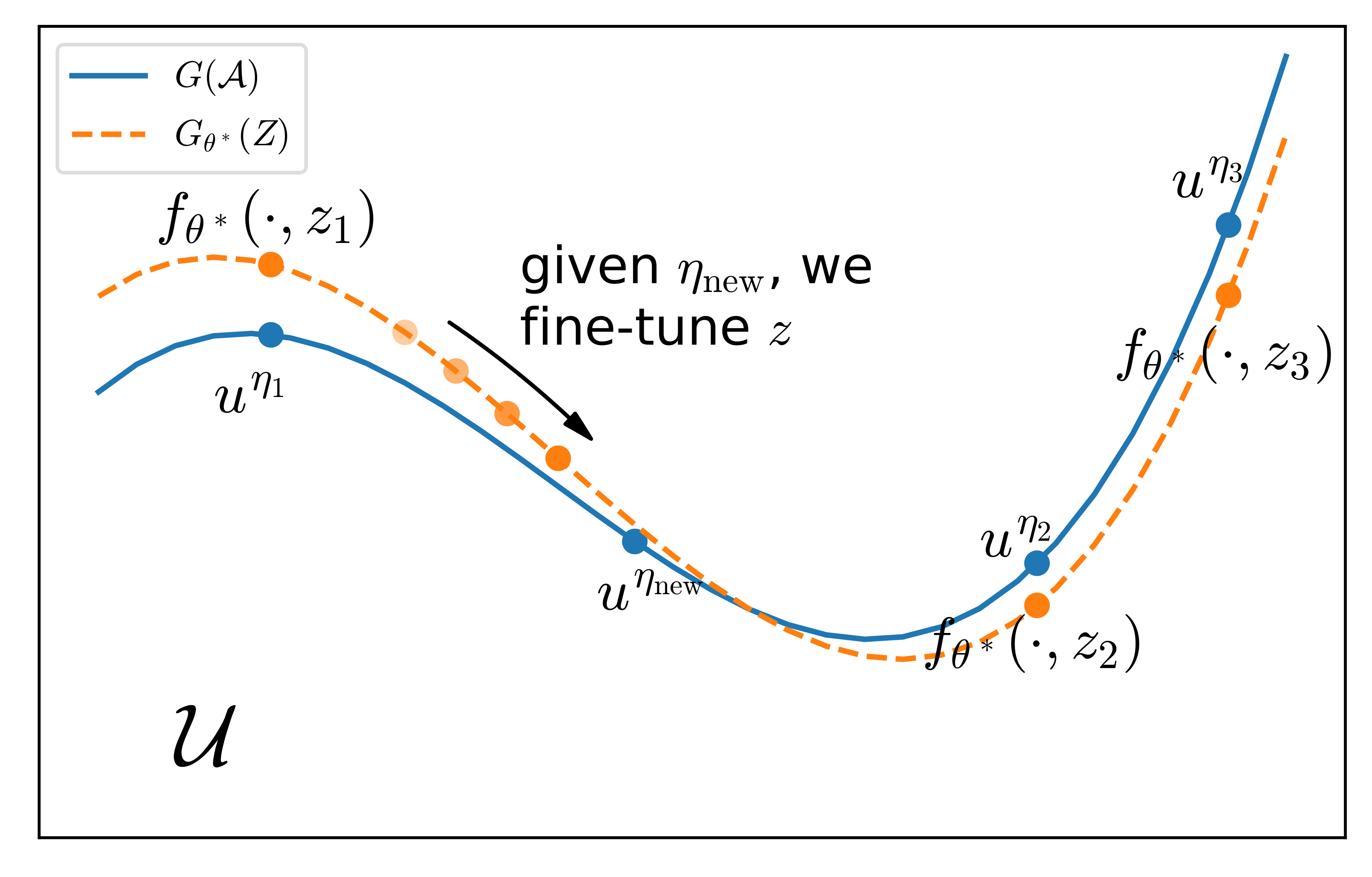}
}
\subfigure[Fine-tune both $z$ and $\theta$ (\textit{MAD-LM})]{
	\label{fig:MADft}
	\includegraphics[width=0.42\columnwidth]{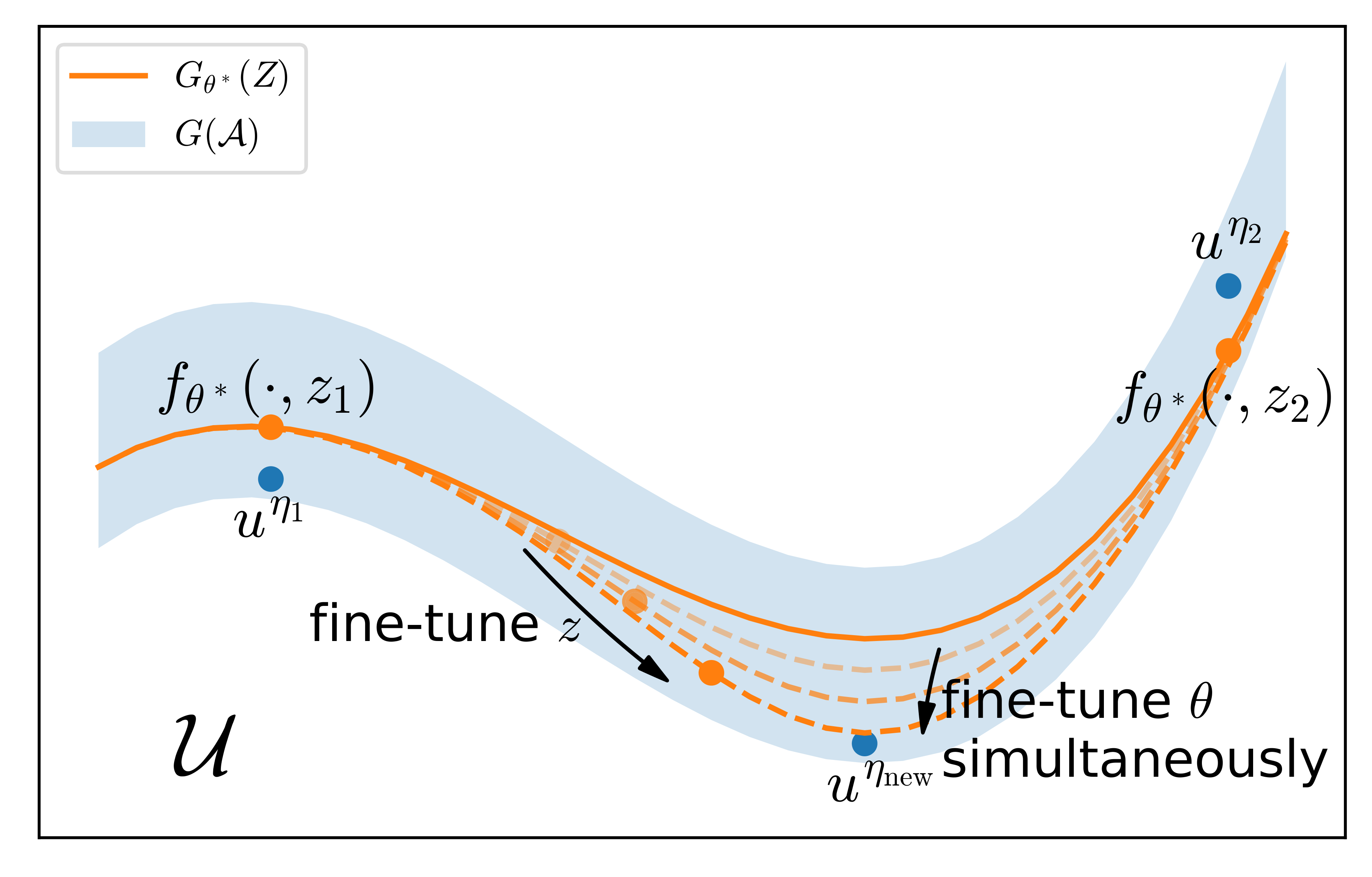}
}
\caption{
	Illustration of how MAD works from the manifold learning perspective. 
	\, \textbf{(a) \textit{MAD-L}}: The function space $\mathcal{U}$ is mapped to a 2-dimensional plane.
	The blue solid curve represents the solution set $G(\mathcal{A})$ formed by exact solutions corresponding to all possible PDE parameters, and each point on the curve represents an exact solution corresponding to one PDE parameter.
	The orange dotted curve represents the solution set $G_{\theta^*}(Z)$ obtained by the pre-trained model, and each point on the curve corresponds to a latent vector $z$.
	Given $\eta_{\text{new}} \in \mathcal{A}$, rather than searching in the entire function space $\mathcal{U}$, \textit{MAD-L} only searches on the orange dotted curve to find an optimal $z$ such that its corresponding solution $u_{\theta^*}(\cdot,z)$ is nearest to the blue point $u^{\eta_{\text{new}}}$.
	\, \textbf{(b) \textit{MAD-LM}}: The solution set $G(\mathcal{A})$ lies within a neighborhood of $G_{\theta^*}(Z)$ that is represented by a gray shadow band. 
	To find solution $u^{\eta_{\text{new}}}$, we have to fine-tune $\theta$ (i.e., the orange dotted lines) and the latent vector $z$ (i.e., the points on the orange dotted lines) simultaneously to approach the exact solution. 
	As the search scope is limited to a strip with a small width, the fine-tuning process 
	can be expected to converge quickly. 
	}
\end{figure*}

We interpret how the \textit{MAD-L} method works from the manifold learning perspective, which also provides a new interpretation of the DeepSDF algorithm~\cite{park2019deepsdf}.
For the rest of this section, the domain $\Omega$ is fixed and excluded from $\eta$ for simplicity. 
Now, we consider the following scenario. 
\begin{scenario}\label{as:lowdim}
	The set of solutions $G(\mathcal{A})=\{G(\eta)\mid\eta\in\mathcal{A}\}\subset\mathcal{U}$ is contained in a low-dimensional structure. 
	To be more specific, there is a finite-dimensional space $Z=\R^l$ (with $l\ll\dim\mathcal{U}$) and a Lipschitz continuous mapping
	$\bar G:Z\to\mathcal{U}$, such that $G(\mathcal{A})\subseteq\bar G(Z)$. 
	In other words, for any $\eta\in\mathcal{A}$, there exists $z\in Z$ satisfying $\bar G(z)=G(\eta)$. 
\end{scenario}
The mapping $\bar G$ is Lipschitz continuous if and only if there exists some $C>0$ such that $\bigl\|\bar G(z)-\bar G(z')\bigr\|_\mathcal{U}\le C\|z-z'\|$ for all $z,z'\in Z$. 
This Lipschitz continuous constraint excludes highly irregular mappings like space-filling curves.
When $\mathcal{A}$ is a finite-dimensional space and $G$ is Lipschitz continuous, the parametric PDE would fall into this scenario (just take $Z=\mathcal{A}$, $\bar G=G$).
Since $\dim Z\ll\dim\mathcal{U}$ (the latter is usually infinity) holds, we may view the mapping $\bar G$ as some sort of ``decoder'', and $Z$ is the corresponding latent vector space, despite of the fact that there doesn't exist an ``encoder''.
In many cases, $\bar G(Z)\subset\mathcal{U}$ forms an embedded submanifold, and therefore our MAD method can be viewed as a manifold-learning approach.
Once the mapping $\bar G$ is learned as above, then for a given parameter $\eta$, 
searching for the solution $u^\eta$ in the whole space $\mathcal{U}$ is no longer needed. 
Instead, we may focus on the smaller subset $\bar G(Z)$, i.e. the class of functions in $\mathcal{U}$ that is parametrized by $Z$, since $u^\eta=G(\eta)\in\bar G(Z)$ holds for any $\eta\in\mathcal{A}$. 
We then solve the optimization problem
\begin{equation}\label{eq:trueDecoderSol}
z^\eta=\operatorname*{\argmin}_{z}L^\eta[\bar G(z)],
\end{equation} 
and $\bar G(z^\eta)$ is the approximate solution. Assuming that the dimension of $Z$ is chosen (either empirically or through trial and error), 
the aim is to find the mapping $\bar G$. 
Since such a mapping is usually complex and hard to design by hand, we consider the $\theta$-parametrized
\footnote{Two types of parametrization are considered here. The latent vector $z$ parametrizes a point on the manifold $\bar G(Z)$ or $G_\theta(Z)$, and $\theta$ parametrizes the shape of the entire manifold $G_\theta(Z)$.}
version $G_\theta:Z\to\mathcal{U}$, and find the best $\theta$ automatically by solving an optimization problem. 
$G_\theta$ can be constructed in the simple form
\begin{equation}
	G_\theta(z)({\widetilde{x}})=u_\theta({\widetilde{x}},z)
,\end{equation} 
where $u_\theta$ is a neural network whose input is the concatenation of ${\widetilde{x}}\in\R^d$ and $z\in\R^l$. 
The next step is to find the optimal model weight $\theta$ via training, with the target being $G(\mathcal{A})\subseteq G_\theta(Z)$. 
Assuming that the PDE parameters are generated from a probability distribution $\eta\sim p_\mathcal{A}$, then $G(\eta)\in G_\theta(Z)$ holds almost surely if and only if
\begin{equation}\label{eq:l2dist}
	\begin{split}
	d(\theta) =\operatorname*{\mathbb{E}}_{\eta\sim p_\mathcal{A}}\left[d_\mathcal{U}\bigl(u^\eta,G_\theta(Z)\bigr)\right]
			  =\operatorname*{\mathbb{E}}_{\eta\sim p_\mathcal{A}}\left[\min_z\bigl\lVert u^\eta-u_\theta(\cdot,z)\bigr\rVert_\mathcal{U}\right] = 0,
	\end{split}
\end{equation}
which suggests taking $\theta^* = \operatorname*{\arg\min}_{\theta} d(\theta)$. 
In case we do not have direct access to the exact solutions $u^\eta$, the equivalent%
\footnote{Assume that the solution of Eq.\eqref{def:PDE} is unique for all $\eta\in\mathcal{A}$, and $u\in\mathcal{U}$ is the solution if and only if $L^\eta[u]=0$.}
condition
\begin{equation}
	d'(\theta)=\operatorname*{\mathbb{E}}_{\eta\sim p_\mathcal{A}}\left[\min_zL^\eta[u_\theta(\cdot,z)]\right] = 0
\end{equation}
is considered, and 
$d'(\theta)$ becomes the alternative loss to be minimized.
In the specific implementation, the expectation on $\eta\sim p_\mathcal{A}$ is estimated by Monte Carlo samples $\eta_1,\dots,\eta_N$, and the optimal network weight $\theta$ is taken to be
\begin{equation}\label{eq:rawMADtr}
	\theta^*\approx\operatorname*{\arg\min}_{\theta}\frac1N\sum_{i=1}^{N}\min_{z_i}L^{\eta_i}[u_\theta(\cdot,z_i)].
\end{equation}
We further estimate the physics-informed loss $L^\eta$ using Monte Carlo method to obtain Eq.\eqref{eq:MADtr}.
After that, when a new PDE parameter $\eta_{\text{new}} \in \mathcal{A}$ comes, a direct adaptation of Eq.\eqref{eq:trueDecoderSol} would then give rise to the fine-tuning process Eq.\eqref{eq:MADinf}, since $u_{\theta^*}(\cdot,z)=G_{\theta^*}(z)\approx\bar G(z)$ holds. 
An intuitive illustration of how \textit{MAD-L} works from the manifold learning perspective is given in Fig.\ref{fig:MAD}. 

\paragraph{A Visualization Example}\label{src:motivational_example}
\begin{figure*}
\centering
\subfigure[Pre-training]{
	\label{fig:MADtr}
	\includegraphics[width=0.5\columnwidth]{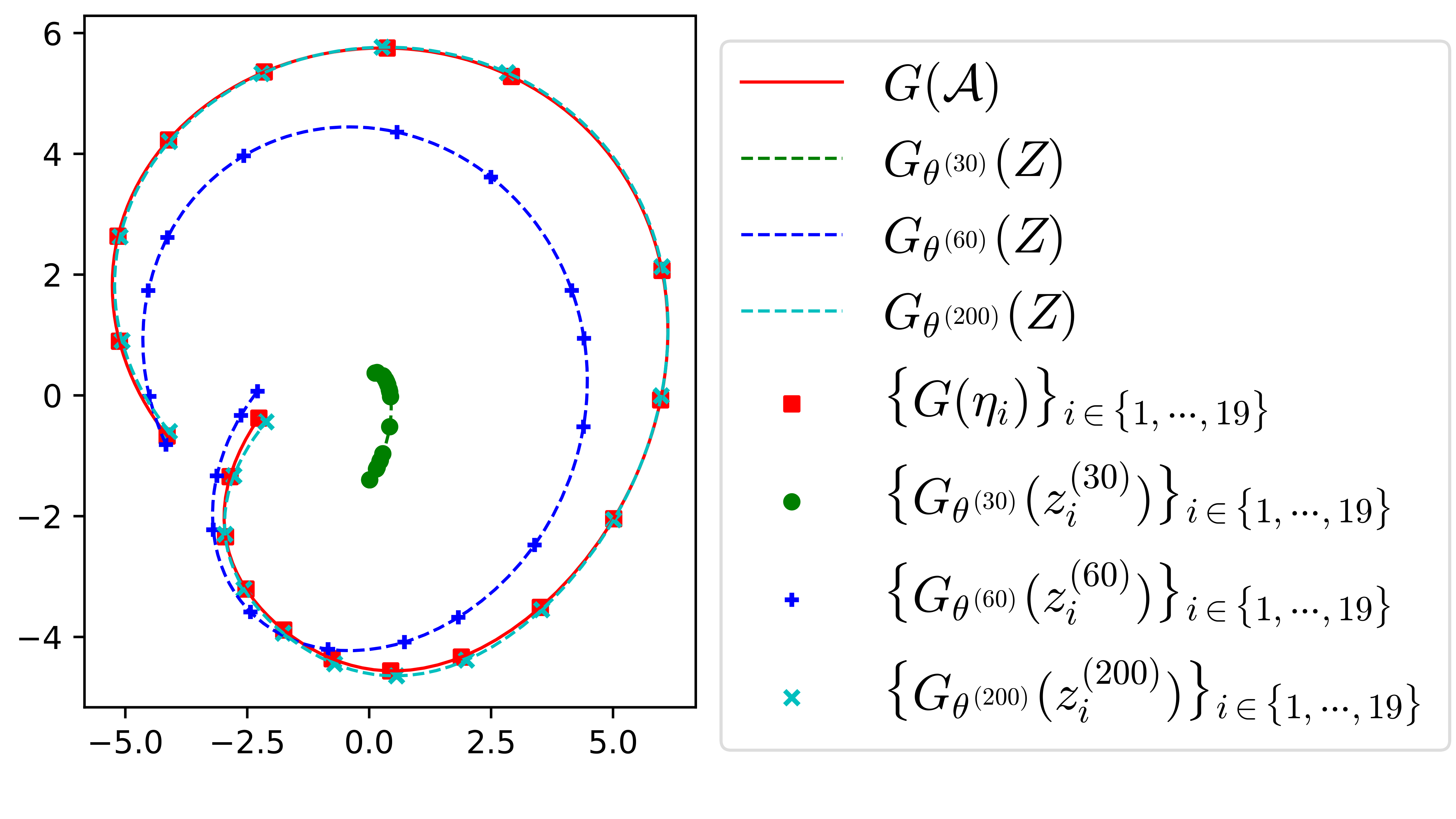}
}
\subfigure[Fine-tuning]{
	\label{fig:MADinfL}
	\includegraphics[width=0.4\columnwidth]{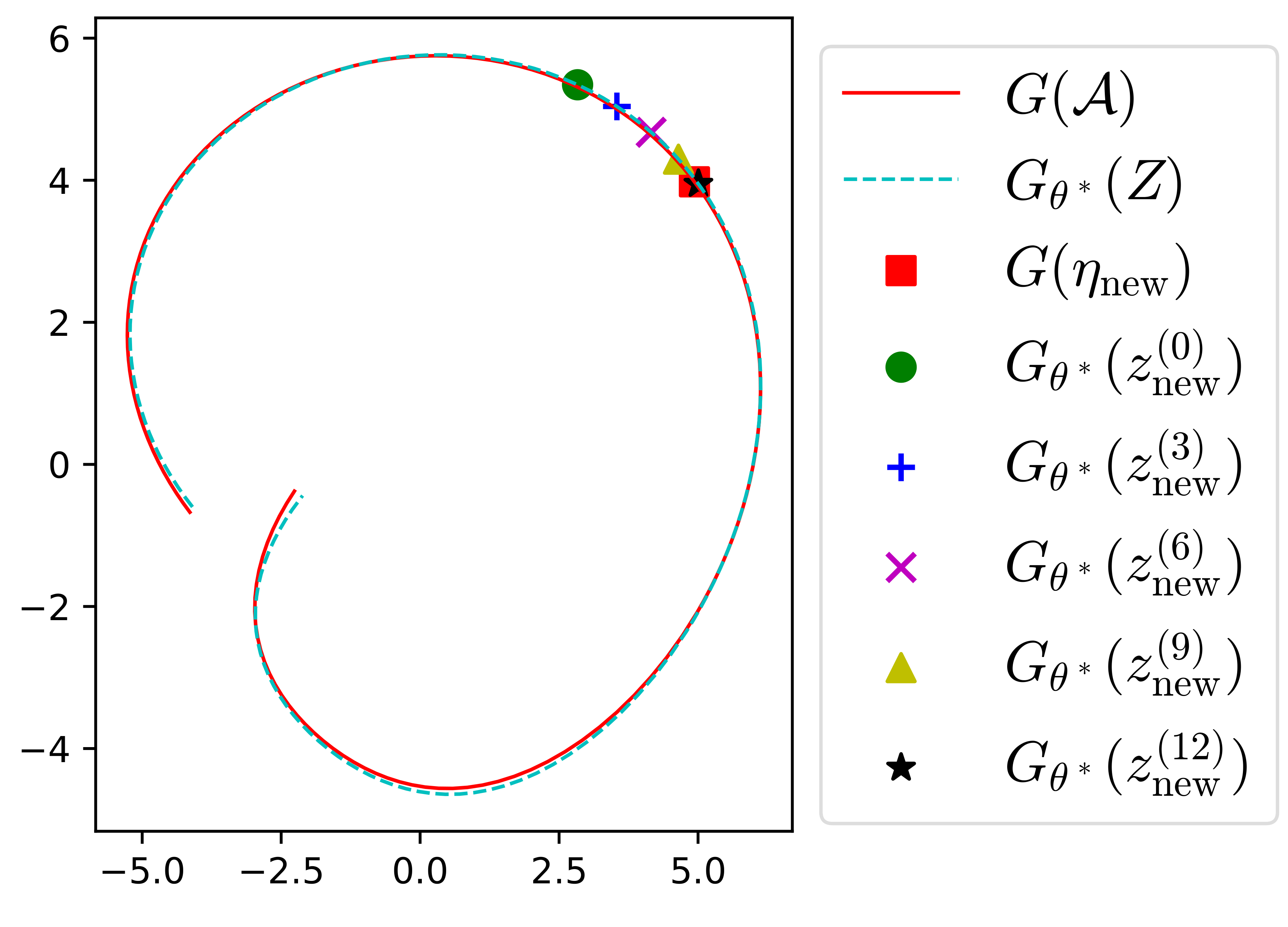}
}
\caption{
Visualization of the MAD pre-training and fine-tuning process for the ODE problem.
}
\label{figs:MAD_example}
\end{figure*}

An ordinary differential equation (ODE) is used to visualize the pre-training and fine-tuning processes of \textit{MAD-L}. Consider the following problem with domain $\Omega=(-\pi,\pi)\subset\R$:
\begin{align}
	\frac{\mathrm{d}u}{\mathrm{d}x}&=2(x-\eta)\cos\bigl((x-\eta)^2\bigr),\qquad u(\pm\pi)=\sin\bigl((\pm\pi-\eta)^2\bigr).
\end{align}
We sample 20 points equidistantly on the interval $[0, 2]$ as variable ODE parameters, and randomly select one $\eta_{\text{new}}$ for fine-tuning stage and the rest $\{\eta_i\}_{i \in \{1,\cdots,19\}}$ for pre-training stage.
\textit{MAD-L} generates a sequence of $(\theta^{(m)}, \{z_i^{(m)}\}_{i\in\{1,\cdots,19\}})$ in pre-training stage, and terminates at $m=200$ with the optimal $(\theta^*, \{z_i^*\}_{i\in\{1,\cdots,19\}})$.
The infinite-dimensional function space $\mathcal{U}=C([-\pi,\pi])$ is projected onto a 2-dimensional plane using Principal Component Analysis (PCA).
Fig.\ref{fig:MADtr} visualizes how $G_\theta(Z)$ gradually fits $G(\mathcal{A})$ in pre-training stage.
The set of exact solutions $G(\mathcal{A})$ forms a 1-dimensional manifold (i.e. the red solid curve), and the marked points $\{G(\eta_i)\}_{i \in \{1,\cdots,19\}}$ represent the corresponding ODE parameters used for pre-training.
Each dotted curve represents a solution set $G_\theta^{(m)}(Z)$ obtained by the neural network at the $m$-th iteration with the points $G_{\theta^{(m)}}(z_i^{(m)})=u_{\theta^{(m)}}(\cdot,z_i^{(m)})$ also marked on the curve.
As the number of iterations $m$ increases, the network weight $\theta=\theta^{(m)}$ updates, making the dotted curves evolve and finally fit the red solid curve, i.e., the target manifold $G(\mathcal{A})$.
Fig.\ref{fig:MADinfL} illustrates the fine-tuning process for a given new ODE parameter $\eta_{\text{new}} \in \mathcal{A}$.
As in Fig.\ref{fig:MADtr}, the red solid curve represents the set of exact solutions $G(\mathcal{A})$, while the cyan dotted curve represents the solution set $G_{\theta^*}(Z) = G_{\theta^{(200)}}(Z)$ obtained by the pre-trained network.
As $z=z_{\text{new}}^{(m)}$ updates (i.e., through fine-tuning $z$), the marked point $G_{\theta^*}(z_{\text{new}}^{(m)})$ moves on the cyan dotted curve, and finally converges to the approximate solution $G_{\theta^*}(z_{\text{new}}^*) = G_{\theta^*}(z_{\text{new}}^{(12)}) \approx G(\eta_{\text{new}})$.

\subsection{Manifold Learning Interpretation of \textit{MAD-LM}}\label{sec:MAD_LM}
The \textit{MAD-L} method is designed for Scenario~\ref{as:lowdim}. 
However, many parametric PDEs encountered in real applications do not fall into this scenario, especially when the parameter set $\mathcal{A}$ of PDEs is an infinite-dimensional function space.
Simply applying \textit{MAD-L} method to these PDE solving problems would likely lead to unsatisfactory results. 
However, \textit{MAD-LM} works in a more general scenario, and thus has the potential of getting improved performance for a wide range of parametric PDE problems. 
This alternative scenario is given as follows. 
\begin{scenario}\label{as:lowdimApprox}
	The solution set $G(\mathcal{A})\subset\mathcal{U}$ can be approximated by a set with low-dimensional structure, in the sense that there is a finite-dimensional space $Z=\R^l$ (with $l\ll\dim\mathcal{U}$) and a Lipschitz continuous mapping $\bar G:Z\to\mathcal{U}$,
	such that $G(\mathcal{A})$ is contained in the $c$-neighborhood of $\bar G(Z)\subset\mathcal{U}$, where $c$ is a relatively small constant. 
	In other words, for any $\eta\in\mathcal{A}$, there exists some $z\in Z$ satisfying $\|\bar G(z)-G(\eta)\|_\mathcal{U}\le c$. 
\end{scenario}

Appendix \ref{appendix:example_assumption2} gives an example of $G(\mathcal{A})$ that falls into Scenario~\ref{as:lowdimApprox} but not Scenario~\ref{as:lowdim}. 
In this new scenario, similar derivation leads to the same pre-training stage, which is used to find the initial decoder mapping $G_{\theta^*}\approx\bar G$. 
However, in the fine-tuning stage, simply fine-tuning the latent vector $z$ won't give a satisfactory solution in general due to the existence of the $c$-gap. 
Therefore, we have to fine-tune the model weight $\theta$ with the latent vector $z$ simultaneously, and solve the optimization problem Eq.\eqref{eq:MADinf_LM}.
It produces a new decoder $G_{\theta_{\text{new}}^*}(z_{\text{new}}^*)$ specific to the parameter $\eta_{\text{new}}$.
An intuitive illustration is given in Fig.\ref{fig:MADft}.


\section{Numerical Experiments}\label{sec:numerical_experiments}
To evaluate the effectiveness of the MAD method, we apply it to solve three parametric PDEs:
(1) Burgers' equation with variable initial conditions;
(2) Maxwell's equations with variable equation coefficients;
and (3) Laplace's equation with variable solution domains and boundary conditions (heterogeneous PDE parameters).
Accuracy of the model is measured by $average\ relative\ L_2\ error$ (abbreviated as $L_2\ error$) between predicted solutions and reference solutions, and we provide the mean value and the 95\% confidence interval of $L_2\ error$.
We compare MAD with other methods including learning from scratch (abbreviated as \textit{From-Scratch}), \textit{Transfer-Learning}~\cite{weinan2018deep}, \textit{MAML}~\cite{finn2017model, antoniou2019train}, \textit{Reptile}~\cite{nichol2018reptile} and \textit{PI-DeepONet}~\cite{wang2021learning}.
For each experiment, the PDE parameters are divided into two sets: $S_1$ and $S_2$.
Parameters in $S_1$ correspond to sample tasks for pre-training, and parameters in $S_2$ correspond to new tasks for fine-tuning. 
See Appendix \ref{appendix:exp_config} for the default experimental setup, and more detailed experimental setups and results for Burgers' equation, Maxwell's equation and Laplace's equation are given in Appendix \ref{appendix:burgers}, \ref{appendix:maxwell} and \ref{appendix:laplace} respectively.
Unless otherwise specified, all the experiments are conducted under the MindSpore\footnote{https://www.mindspore.cn/}.

\begin{figure*}
\centering
\subfigure[Burgers' equation]{
    \label{fig:burgers_compare_other_method}
    \includegraphics[width=0.315\columnwidth]{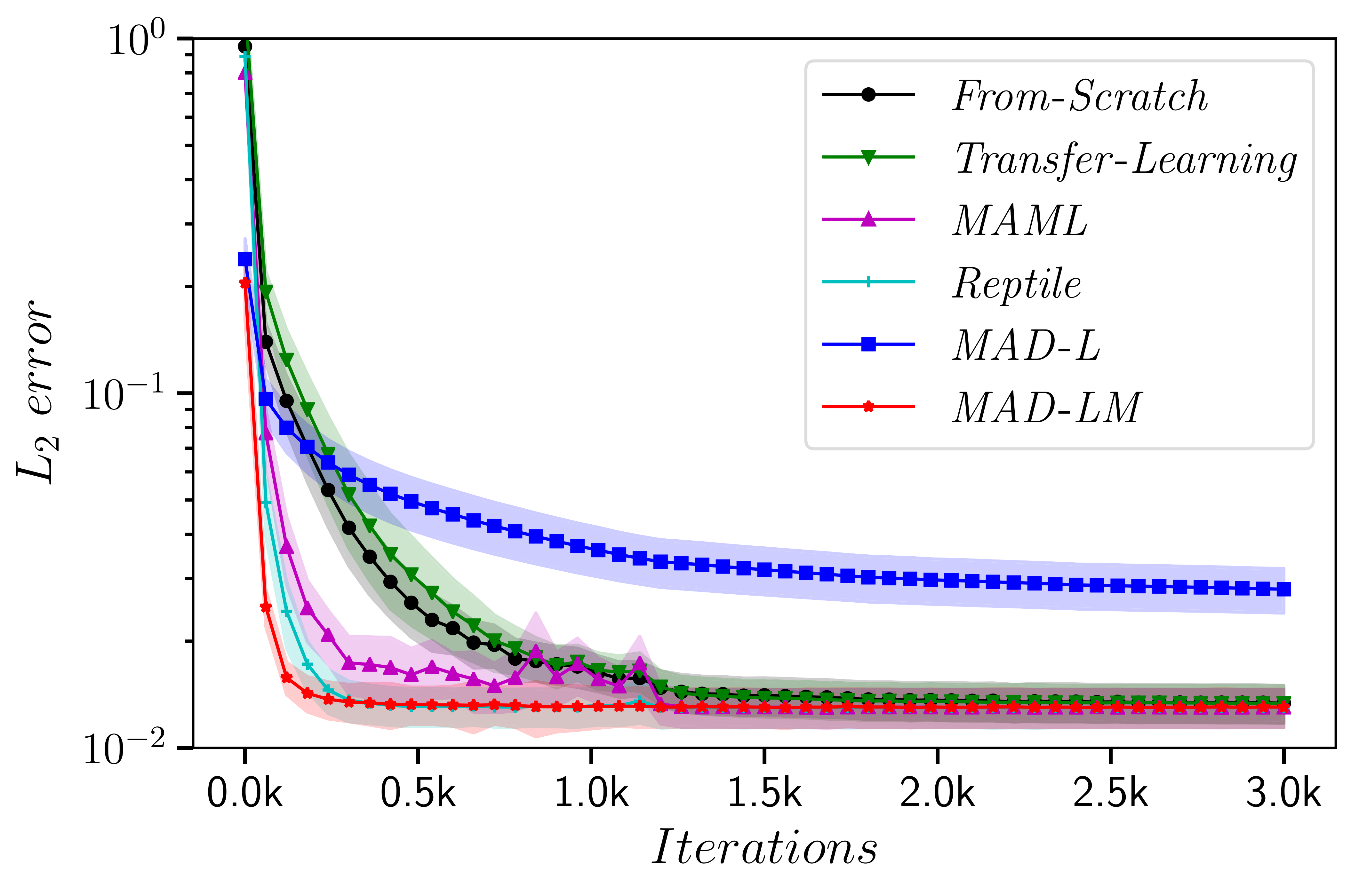}
}
\subfigure[Maxwell's equations]{
    \label{fig:maxwell_compare_other_method}
    \includegraphics[width=0.315\columnwidth]{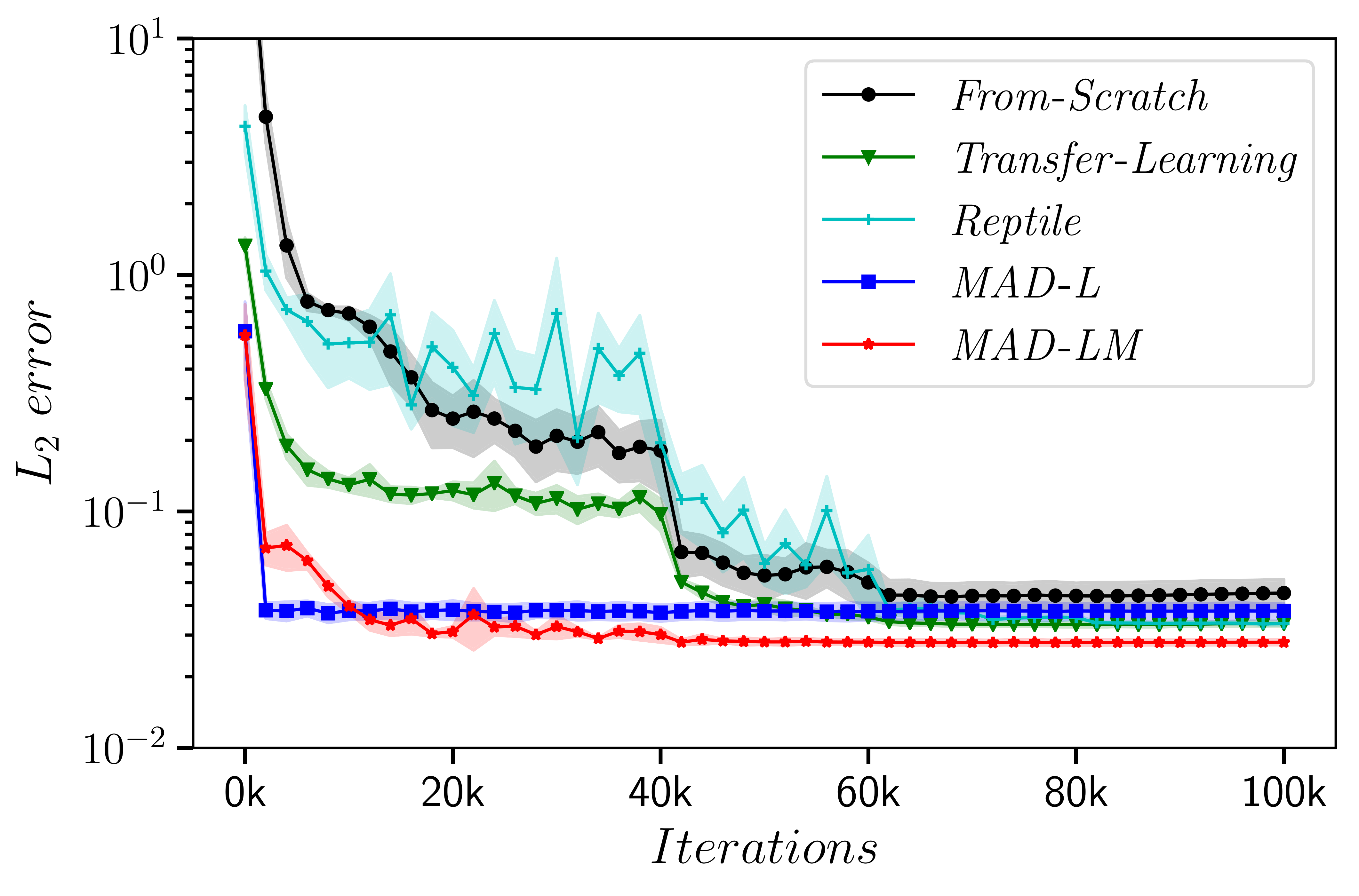}
}
\subfigure[Laplace's equation]{
    \label{fig:laplace_compare_other_method}
    \includegraphics[width=0.315\columnwidth]{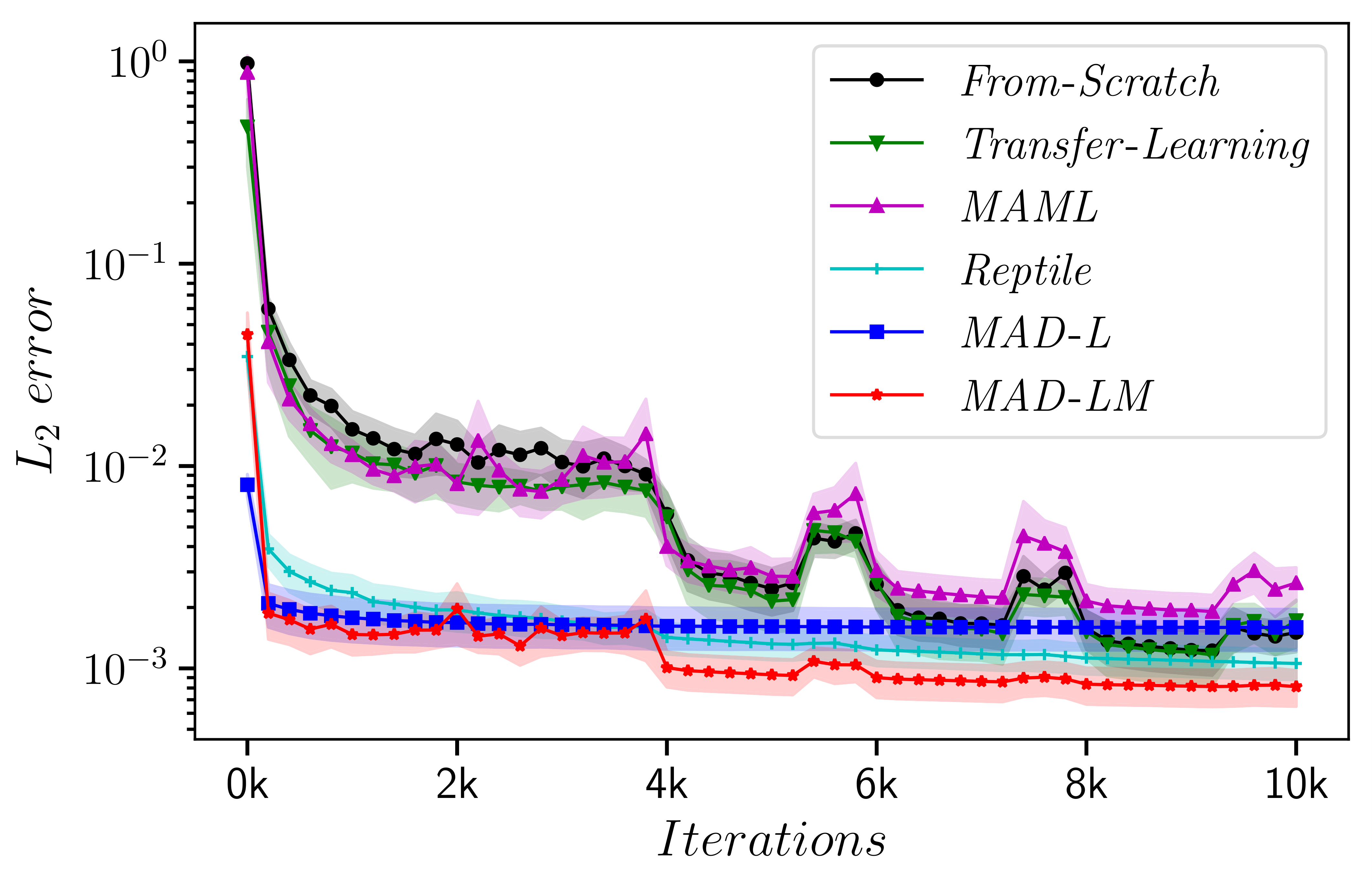}
}
\caption{
    The mean $L_2\ error$ convergence with respect to the number of training iterations.
}
\end{figure*}

\subsection{Burgers' Equation}\label{sec:burgers}

We consider the 1-D Burgers' equation:
\begin{equation}\label{def:burgers}
\begin{aligned}
\frac{\partial u}{\partial t} + u \frac{\partial u}{\partial x} = \nu \frac{\partial ^2 u}{\partial x^2}, \; x \in (0,1), t \in (0,1],\qquad 
u(x,0) = u_0(x), \; x \in (0,1),
\end{aligned}
\end{equation}
Eq.\eqref{def:burgers} can model one-dimensional flow of a viscous fluid, where $u$ is the velocity, $\nu$ is the viscosity coefficient and initial condition $u_0(x)$ is the changing parameter of the PDE, i.e. $\eta=u_0(x)$. 
The initial condition $u_0(x)$ is generated using Gaussian random field (GRF) \cite{liu2019advances} according to $u_0(x) \sim \mathcal{N}(0; 100(- \Delta + 9I)^{-3})$ with periodic boundary conditions.

Fig.\ref{fig:burgers_compare_other_method} shows the mean $L_2\ error$ of all methods as the number of training iterations increases. 
All methods converge to nearly the same accuracy (the mean $L_2\ error$ close to 0.013) except for \textit{MAD-L}, which we guess probably due to the $c$-gap introduced in Sec.\ref{sec:MAD_LM}. 
In terms of convergence speed, \textit{From-Scratch} and \textit{Transfer-Learning} need about 1200 iterations to converge, whereas \textit{MAML}, \textit{Reptile} and \textit{MAD-LM} need about 200 iterations to converge. 
\textit{MAD-LM} has the fastest convergence speed, requiring only 17\% of the training iterations of \textit{From-Scratch}. 
In this experiment, \textit{Transfer-Learning} does not show any advantage over \textit{From-Scratch}, which means that \textit{Transfer-Learning} fails to obtain any useful knowledge in pre-training stage.

\begin{table}
\caption{The mean $L_2\ error$ of \textit{PI-DeepONet} and MAD under different numbers of samples in $S_1$.}
\label{tb:ad_vs_pi_deeponet}
\centering
\begin{tabular}{cccc}
    \toprule
    $|S_1|$ & \textit{PI-DeepONet} & \textit{MAD-L} & \textit{MAD-LM} \\
    \midrule
    10  &  0.715  & 0.365 & 0.015 \\
    50  &  0.247  & 0.046 & 0.013 \\
    100 &  0.217  & 0.028 & 0.013 \\
    200 &  0.169  & 0.020 & 0.013 \\
    300 &  0.181  & 0.018 & 0.013 \\
    400 &  0.183  & 0.019 & 0.013 \\
    \bottomrule
\end{tabular}
\end{table}

\textit{PI-DeepONet} can directly inference for unseen PDE parameters in $S_2$, so it has no fine-tuning process.
Table \ref{tb:ad_vs_pi_deeponet} shows the comparison of the mean $L_2\ error$ of \textit{PI-DeepONet} and MAD under different numbers of training samples in $S_1$.
The results show that \textit{PI-DeepONet} has a strong dependence on the number of training samples, and its mean $L_2\ error$ is remarkably high when $S_1$ is small. Moreover, its mean $L_2\ error$ is significantly higher than that of \textit{MAD-L} or \textit{MAD-LM} in all cases.

\begin{figure*}
\centering
\subfigure[Burgers' equation]{
    \label{fig:burgers_extrapolation_grf}
    \includegraphics[width=0.35\columnwidth]{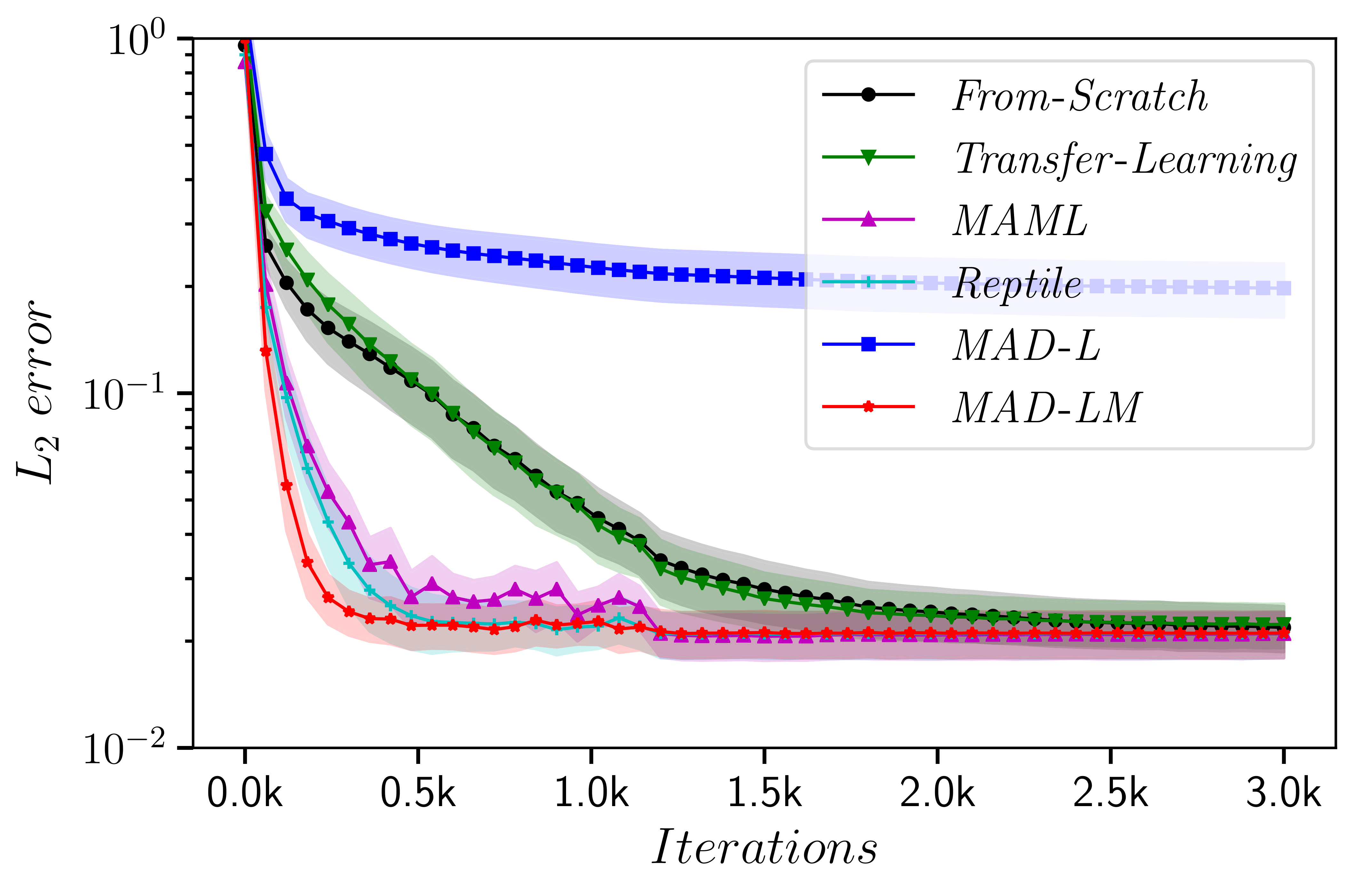}
}
\subfigure[Maxwell's equations]{
    \label{fig:maxwell_extrapolation}
    \includegraphics[width=0.35\columnwidth]{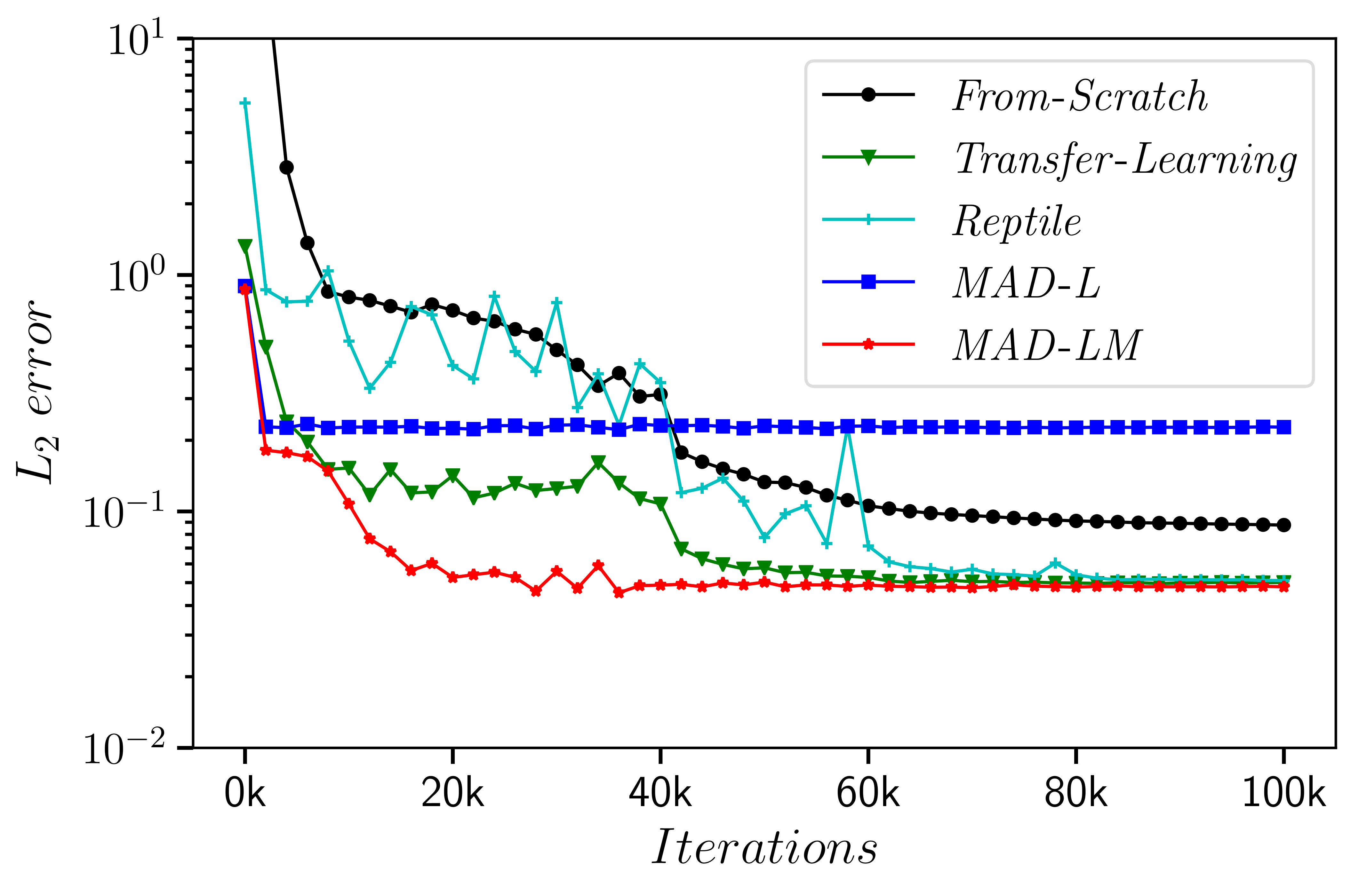}
}
\caption{
    The mean $L_2\ error$ convergence with respect to the number of training iterations for extrapolation experiments.
}
\end{figure*}

In the above experiments, $\eta$s in $S_1$ and $S_2$ come from the same GRF, so we can assume that the tasks in the pre-training stage come from the same task distribution as the tasks in the fine-tuing stage. 
We investigate the extrapolation capability of MAD, that is, tasks in the fine-tuing stage come from different task distribution than those in the pre-training stage.
Specifically, $S_1$ is still the same as above, but $S_2$ is generated from $\mathcal{N}(0; 100(- \Delta + 25I)^{-2.5})$.
Fig.\ref{fig:burgers_extrapolation_grf} shows the results of extrapolation experiments.
Since the distribution of tasks has changed, the manifold learned in the pre-training stage fits $G(\mathcal{A})$ worse, so \textit{MAD-L} exhibits worse accuracy than that in Fig.\ref{fig:burgers_compare_other_method}.
However, as with Fig.\ref{fig:burgers_compare_other_method}, the convergence speed of \textit{MAD-LM} in Fig.\ref{fig:burgers_extrapolation_grf} is also better than other methods.
This shows that the extrapolation capability of MAD is also better than other methods in this example.

\subsection{Time-Domain Maxwell's Equations}\label{sec:maxwell}

We consider the time-domain 2-D Maxwell's equations with a point source in the transverse Electric (TE) mode \cite{gedney2011introduction}:
\begin{equation}\label{eq:point_src}
\begin{aligned}
\frac{\partial E_x}{\partial t } = \frac{1}{\epsilon_0\epsilon_r} \frac{\partial H_z}{\partial y}, \quad
\frac{\partial E_y}{\partial t } = -\frac{1}{\epsilon_0\epsilon_r} \frac{\partial H_z}{\partial x}, \quad
\frac{\partial H_z}{\partial t } = -\frac{1}{\mu_0\mu_r} \left(\frac{\partial E_y}{\partial x} - \frac{\partial E_x}{\partial y} + J\right),
\end{aligned}
\end{equation}
where $E_x$, $E_y$ and $H_z$ are the electromagnetic fields, $J$ is the point source term.
The equation coefficients $\epsilon_0$ and $\mu_0$ are the permittivity and permeability in vacuum, respectively.
The equation coefficients $\epsilon_r$ and $\mu_r$ are the relative permittivity and relative permeability of the media, respectively.
\cite{huang2021solving} uses modified PINNs method to solve Eq.\eqref{eq:point_src} with fixed $\epsilon_r=1$ and $\mu_r=1$.
However, in this paper, $(\epsilon_r$, $\mu_r)$ are variable parameters of the PDEs, i.e., $\eta=(\epsilon_r$, $\mu_r)$, which corresponds to the media properties in the simulation region. 

Fig.\ref{fig:maxwell_compare_other_method} shows that all methods converge to similar accuracy (mean $L_2\ error$ is close to 0.04), and \textit{MAD-LM} achieves the lowest mean $L_2\ error$ (0.028). 
In terms of convergence speed, \textit{MAD-L} and \textit{MAD-LM} are obviously superior to other methods. 
It is worth noting that \textit{MAML} fails to converge in pre-training stage, therefore its data is missing in Fig.\ref{fig:maxwell_compare_other_method}.
We guess the reason is that singularity brought by point source and computation of second-order derivatives pose great difficulties in solving optimization problem. 
\textit{Reptile} also does not show good generalization ability probably due to the same singularity problem.

We do an extrapolation experiment, where $(\epsilon_r, \mu_r)$ in $S_1$ comes from $[1,5]^2$, but in the fine-tuning stage, we only consider the $(\epsilon_r, \mu_r)=(7, 7)$ case.
Because the extrapolated task does not lie in the task distribution in the pre-training stage, the point $G(\eta_\text{new})$ corresponding to the extrapolated task in the function space $\mathcal{U}$ is not on the learned manifold $G_{\theta^*}(Z)$, which causes \textit{MAD-L} to converge to poor accuracy.
Fig.\ref{fig:maxwell_extrapolation} indicates that \textit{MAD-LM} is significantly faster than \textit{From-Scratch} and \textit{Reptile} in convergence speed while maintaining high accuracy.
Notably, \textit{Transfer-Learning} also exhibits faster convergence than \textit{From-Scratch} and \textit{Reptile}.
This is because $(\epsilon_r, \mu_r)=(4, 5)$ is randomly selected in the pre-training stage and is very close to $(\epsilon_r, \mu_r)=(7, 7)$ in Euclidean distance.

\subsection{Laplace's Equation}\label{sec:laplace}

We consider the 2-D Laplace's equation as follows:
\begin{equation}\label{def:laplace}
\begin{aligned}
    \frac{\partial ^2 u}{\partial x^2}  + \frac{\partial ^2 u}{\partial y^2} = 0, \; (x,y) \in \Omega, \qquad
    u(x,y) &= g(x,y), \; (x,y) \in \partial \Omega,
\end{aligned}
\end{equation}
where the shape of $\Omega$ and boundary condition $g(x,y)$ are the variable parameters of the PDE, i.e. $\eta=(\Omega, g(x,y))$.
In this experiment, we use triangular domain $\Omega$ and vary the shape of $\Omega$ by randomly choosing three points on the circumference of a unit circle to form the triangle. 
Given that $h$ is the boundary condition on the unit circle, we use GRF to generate $h \sim \mathcal{N}(0, 10^{3/2}(-\Delta + 100I)^{-3})$ with periodic boundary conditions.
The analytical solution of the Laplace's equation on the unit circle can be obtained by Fourier method. 
Then, we use the analytical solution on the three sides of the triangle as the boundary condition $g(x,y)$.
The variable PDE parameters include the shape of the solution domain (the shape of the triangle) and the boundary conditions on the three sides of the triangle, so the PDE parameters here are heterogeneous.
MAD can implicitly encode such heterogeneous PDE parameters as latent vectors conveniently, whereas \textit{PI-DeepONet} is unable to handle this case without further adaptations.

Fig.\ref{fig:laplace_compare_other_method} shows that all methods finally converge to similar accuracy (mean $L_2\ error$ close to 0.001), and \textit{MAD-LM} achieves the lowest mean $L_2\ error$. 
\textit{MAD-L}, \textit{MAD-LM} and \textit{Reptile} show good generalization capability and excellent convergence speed, whereas \textit{Transfer-Learning} and \textit{MAML} do not show any advantage over \textit{From-Scratch}.



\paragraph{Summary of experimental results.}
Achieving fast adaptation is the major focus of this paper, and solutions within a reasonable precision need to be found. 
Indeed, in many control and inverse problems, a higher precision in solving the forward problem (such as parametric PDEs) does not always lead to better results. 
For example, a solution with about $5\%$ relative error is already enough for Maxwell's equations in certain engineering scenarios. 
We are therefore interested in reducing the cost of solving the PDE with a new set of parameters by using only a relatively small number of iterations, in order to obtain an accurate enough solution in practice. 
The advantages of MAD (especially \textit{MAD-LM}) is directly validated in the numerical experiments, as it achieves very fast convergence in the early stage of the training process. 
Some other applications may focus more on the final precision, and is not as sensitive to the training cost. 
In this alternate criterion, the superiority of the MAD method becomes less obvious in our test cases except Maxwell's equations, but its performance is still comparable to other methods. 

\section{Conclusions}\label{sec:conclusion}
In this paper, a novel mesh-free and unsupervised deep learning method MAD is proposed for solving parametric PDEs based on meta-learning idea.
A good initial model is obtained in pre-training stage to learn useful information from a set of sampled tasks, which is then used to help solve the parametric PDEs quickly in fine-tuning stage. 
Moreover, MAD can implicitly encode heterogeneous PDE parameters as latent vectors.
The effectiveness of MAD method is analyzed from the perspective of manifold learning and verified by extensive numerical experiments.

\section*{Acknowledgments}
This work was supported by National Key R\&D Program of China under Grant No. 2021ZD0110400.

\bibliographystyle{unsrtnat}
\bibliography{references}

\newpage
\appendix

\section{An Example for Scenario \ref{as:lowdimApprox}}
\label{appendix:example_assumption2}


We give an example of $G(\mathcal{A})$ which falls into Scenario~\ref{as:lowdimApprox} but not Scenario~\ref{as:lowdim}. 
Let $\{\phi_k(x)\}_{k\in\mathbb{N}}$ be an orthonormal basis of a Hilbert space $L^2(X)$, and $(\lambda_k)_{k\in\mathbb{N}}$ be a sequence of positive real numbers with $\sum_{k=1}^{\infty}\lambda_k<\infty$. 
We take
\begin{equation}
	\mathcal{A}=\Biggl\{\sum_{k=1}^{\infty}\xi_k\sqrt{\lambda_k}\phi_k(x)\ \Biggm|\ \xi_k\in[-1,1]\Biggr\}\subset L^2(X)
,\end{equation}
and consider the equation $u_t=-u$ on $\Omega=X\times[0,1]$ with initial condition $u(x,0)=\eta(x)$. 
It is easy to see that
\begin{equation}
	G(\mathcal{A})=\Biggl\{\sum_{k=1}^{\infty}\xi_k\sqrt{\lambda_k}\phi_k(x)e^{-t}\ \Biggm|\ \xi_k\in[-1,1]\Biggr\}\subset \mathcal{U}=L^2(\Omega)
.\end{equation}
and no $(Z,\bar G)$-pair can make Scenario~\ref{as:lowdim} valid as $G(\mathcal{A})$ is not finite-dimensional. 
As for Scenario~\ref{as:lowdimApprox}, once the number $c$ is given, there exists a large enough $l$ satisfying $\sum_{k=l+1}^{\infty}\lambda_k\le c^2$. 
Then we let $Z=\R^l$ and choose a linear mapping $\bar G$ such that
\begin{equation}
	\bar G\bigl((\xi_k)_{k=1}^l\bigr)=\sum_{k=1}^{l}\xi_k\sqrt{\lambda_k}\phi_k(x)e^{-t}
.\end{equation}
For any $\eta=\sum_{k=1}^{\infty}\xi_k\sqrt{\lambda_k}\phi_k(x)\in\mathcal{A}$, taking $z=(\xi_k)_{k=1}^l\in Z$ gives
\[\begin{split}
	\left\|\bar G(z)-G(\eta)\right\|_{\mathcal{U}}^2 &= \left\|\sum_{k=l+1}^{\infty}\xi_k\sqrt{\lambda_k}\phi_k(x)e^{-t}\right\|_{L^2(X\times[0,1])}^2
	\\&=\sum_{k=l+1}^{\infty}\xi_k^2\lambda_k\int_X\phi_k(x)^2\,\mathrm{d}x\int_0^1e^{-2t}\,\mathrm{d}t
	\\&=\sum_{k=l+1}^{\infty}\xi_k^2\lambda_k\frac{1-e^{-2}}{2}
	\\&\le\sum_{k=l+1}^{\infty}\lambda_k
	\\&\le c^2
.\end{split}\]
This indicates that Scenario~\ref{as:lowdimApprox} is indeed more general than Scenario~\ref{as:lowdim}.

\section{Default Experimental Configurations}\label{appendix:exp_config}
Below is a detailed explanation of the comparative methods covered in the paper.
\begin{itemize}
	\item \textit{From-Scratch}: Train the model from scratch based on the PINNs method for all PDE parameters in $S_2$, case-by-case.
	\item \textit{Transfer-Learning} \cite{weinan2018deep}: Randomly select a PDE parameter in $S_1$ for pre-training stage based on the PINNs method, and then load the obtained weight in pre-training stage for PDE parameters in $S_2$ during fine-tuning stage. 
	\item \textit{MAML} \cite{finn2017model, antoniou2019train}: Meta-train the model for all PDE parameters in $S_1$ based on MAML algorithm.
	In the meta-testing stage, we load the pre-trained weight $\theta^*$ and fine-tune the model for each PDE parameter in $S_2$.
	\item \textit{Reptile} \cite{nichol2018reptile}: Similar to \textit{MAML}, except that the model weight is updated using the Reptile algorithm in the meta-training stage.
 	\item \textit{PI-DeepONet} \cite{wang2021learning}: The model is trained based on the training method proposed in \cite{wang2021learning} for all PDE parameters in $S_1$, and the inference is performed directly for the parameters in $S_2$.
	\item \textit{MAD-L}: Pre-train the model for all PDE parameters in $S_1$ based on our proposed method and then load and freeze the pre-trained weight $\theta^*$ for the second stage.
	In the fine-tuning stage, we choose a $z_i^*$ obtained in the pre-training stage to initialize a latent vector for each PDE parameter in $S_2$, and fine-tune the latent vector.
	The selection of $z_i^*$ is based on the distance between $\eta_{\text{new}}$ and $\eta_i$.
	\item \textit{MAD-LM}: Different from \textit{MAD-L} that freezes the pre-trained weight, we fine-tune the model weight $\theta$ and the latent vector $z$ simultaneously in the fine-tuning stage.
\end{itemize}

Unless otherwise specified, the following default configurations are used for the experiments:
\begin{itemize}
	\item In each iteration, the batch sizes are selected as ($M_{r}$, $M_{bc}$) = (8192, 1024).
	\item For fairness of comparison, the network architectures of all methods (excluding PI-DeepONet) involved in comparison are the same except for the input layer due to the latent vector. 
    For Burgers' equation and Laplace's equation, the standard fully-connected neural networks with 7 fully-connected layers and 128 neurons per hidden layer are taken as a default network structure.
	For Maxwell's equations, the MS-SIREN network architecture \cite{huang2021solving} is used that has 4 subnets, each subnet has 7 fully-connected layers and 64 neurons per hidden layer.
	It is worth noting that our proposed method has gains in different network architectures, and we choose the default network architecture that can achieve high accuracy for the \textit{From-Scratch} method to conduct our comparative experiments.
	\item The network architecture of \textit{PI-DeepONet} used for Burgers' equation is such that both branch net and trunk net are 7 fully-connected layers and 128 neurons per hidden layer.
	\item Sine function \cite{sitzmann2020implicit} is used as the activation function, as it exhibits better performance than other alternatives such as ReLU and Tanh.
	\item The dimension of the latent vector $z$ is determined by trial and error and is set to 128 for Burgers' equation and Laplace's equation, and 16 for Maxwell's equations.
	\item The Adam optimizer \cite{kingma2014adam} is used with the initial learning rate set to 1e-3 or 1e-4 (whoever achieves the best performance). 
	When the training process reaches 40\%, 60\% and 80\%, the learning rate is attenuated by half.
\end{itemize}


\section{Detailed Experimental Setup and Extended Results for Burgers' Equation}\label{appendix:burgers}
We set the viscosity to $\nu = 0.01$ and solve the Eq.\eqref{def:burgers} using open source code implemented by \cite{li2020fourier} to generate the reference solutions.
The spatiotemporal mesh size of the ground truth is $1024\times101$.
In order to solve the Eq.\eqref{def:burgers} better by using the PINNs-based method, we use the hard constraint on periodic boundary condition mentioned in \cite{lu2021physics}.


We generated 150 different initializations of  $u_0(x)$ using GRF and randomly selected 100 cases  as $S_1$. 
The remaining 50 cases are used  as $S_2$ for fine-tuning. 
For \textit{MAD-L} and \textit{MAD-LM}, the pre-training stages run for $50$k iterations while the \textit{Transfer-Learning} pre-trains 3k steps since it only handles one single case. 

\begin{figure}
\begin{center}
\centerline{\includegraphics[width=0.8\columnwidth]{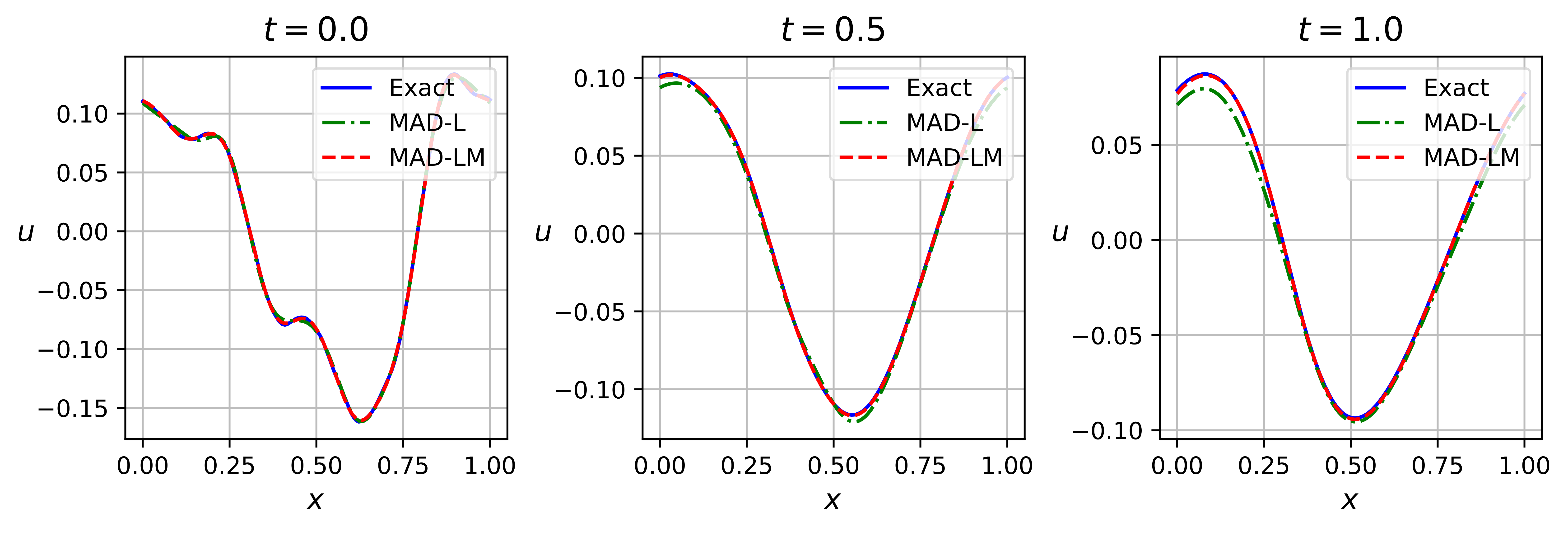}}
\caption{\textbf{Burgers' equation:} Reference solutions vs. model predictions at $t = 0.0$, $t = 0.5$ and $t = 1.0$, respectively.}
\label{fig:burgers_result}
\end{center}
\end{figure}
Fig.\ref{fig:burgers_result} shows model predictions of \textit{MAD-L} and \textit{MAD-LM} compared with the reference solutions under a randomly selected $u_0(x)$ in $S_2$. 
The predictions of \textit{MAD-L} are in overall approximate agreement with the reference solutions, but the fit is poor at the spikes and troughs.
However, the prediction results of \textit{MAD-LM} is almost the same as that of the reference solutions.

\begin{figure}
\centering
\subfigure[\textit{MAD-L}]{
    \label{fig:burgers_ablation_samples_MAD_l}
    \includegraphics[width=0.45\columnwidth]{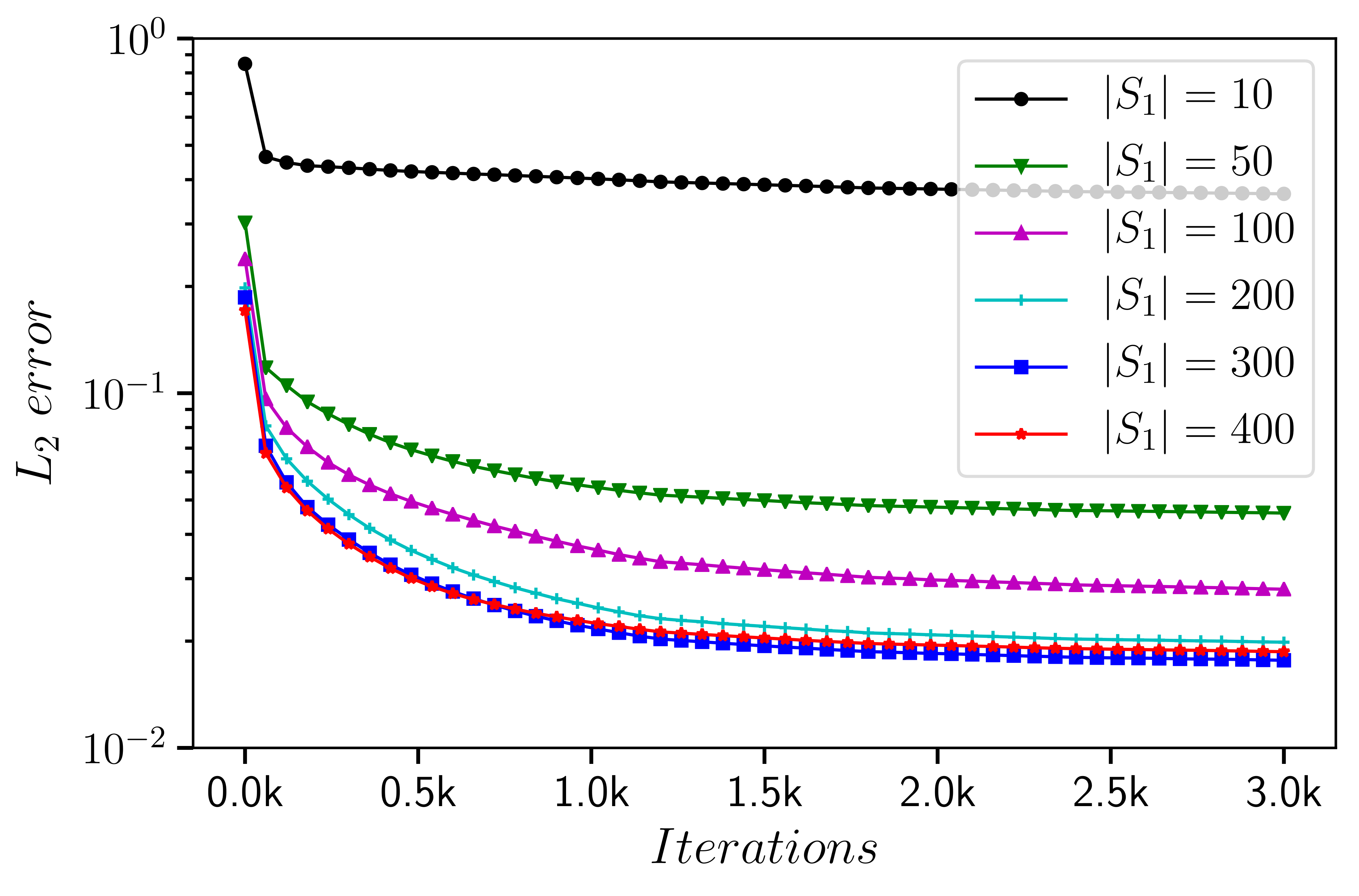}
}
\subfigure[\textit{MAD-LM}]{
    \label{fig:burgers_ablation_samples_MAD_lm}
    \includegraphics[width=0.45\columnwidth]{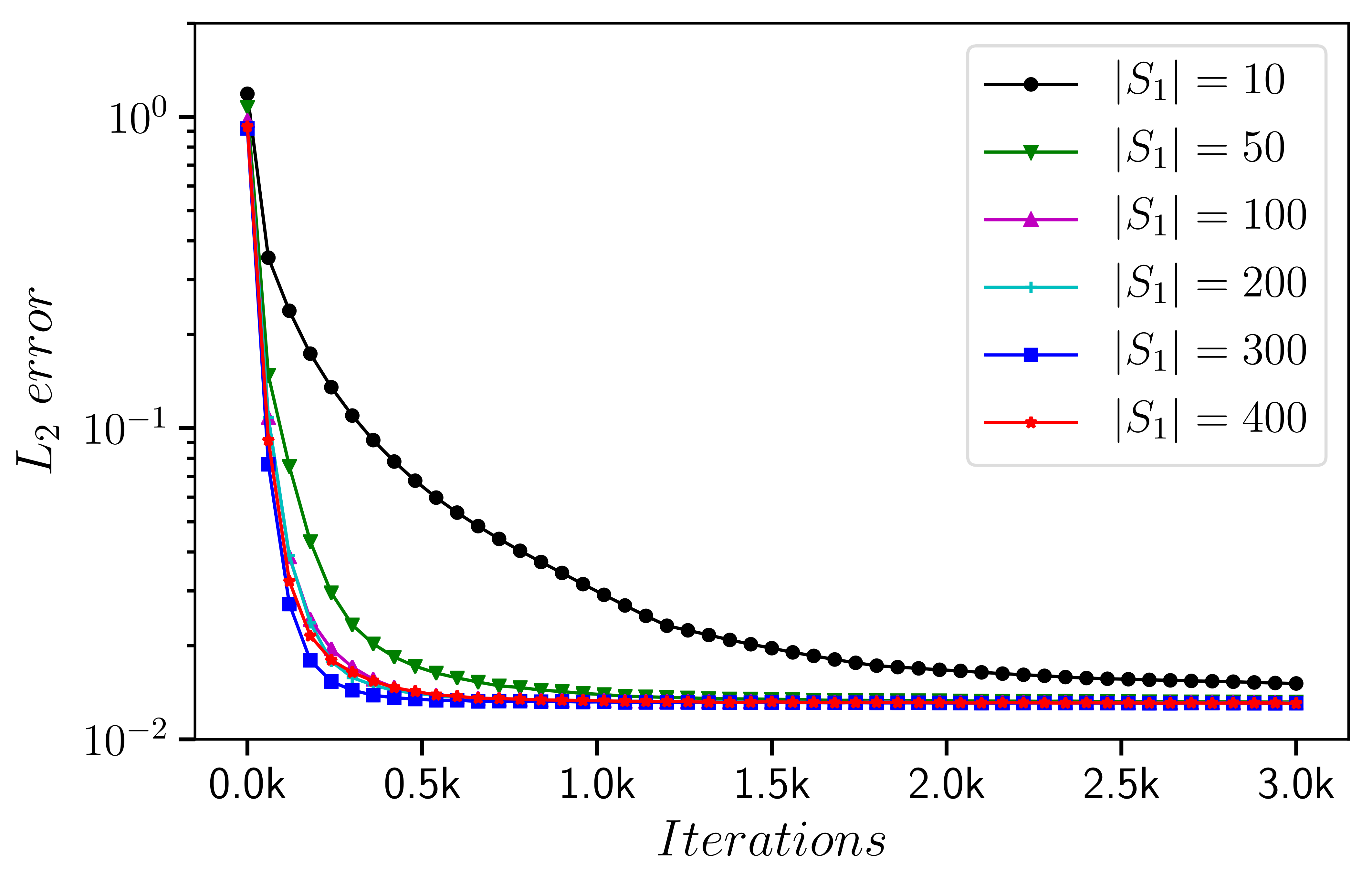}
}
\caption{\textbf{Burgers' equation:} The mean $L_2\ error$ of \textit{MAD-L} (a) and \textit{MAD-LM} (b) convergence with respect to the number of training iterations under different numbers of samples in $S_1$.}
\end{figure}
We investigated the effect of  the number of samples $|S_1|$ in the pre-training stage on \textit{MAD-L} and \textit{MAD-LM}. 
Fig.\ref{fig:burgers_ablation_samples_MAD_l} shows that the accuracy of \textit{MAD-L} after convergence increases with $|S_1|$.
However, when $|S_1|$ reaches about 200, increasing $|S_1|$ does not improve the accuracy of the \textit{MAD-L}.
This result is consistent with the phenomenon shown in Fig.\ref{fig:MADft}.
More samples in the pre-training stage allow the $G_{\theta^*}(Z)$ to gradually fall within the region formed by $G(\mathcal{A})$.
However, when $|S_1|$ reaches a certain level, the $G_{\theta^*}(Z)$ only swings in the region of $G(\mathcal{A})$.
Only optimizing $z$ can make the solution move inside the manifold formed by $G_{\theta^*}(Z)$, but  $u^{\eta_{\text{new}}}$ may not be close enough to the manifold.
Therefore, in order to obtain a more accurate solution, we need to fine-tune $z$ and $\theta$ simultaneously.
Fig.\ref{fig:burgers_ablation_samples_MAD_lm} shows that the accuracy and convergence speed of \textit{MAD-LM} do not change significantly with the increase of samples in the pre-training stage. 
It is only when the number of samples is very small (i.e., $|S_1| = 10$) that the early convergence speed is significantly affected.
This shows that \textit{MAD-LM} can perform well in the fine-tuning stage without requiring a large number of samples during the pre-training stage.

\begin{figure}
\begin{center}
\centerline{\includegraphics[width=0.7\columnwidth]{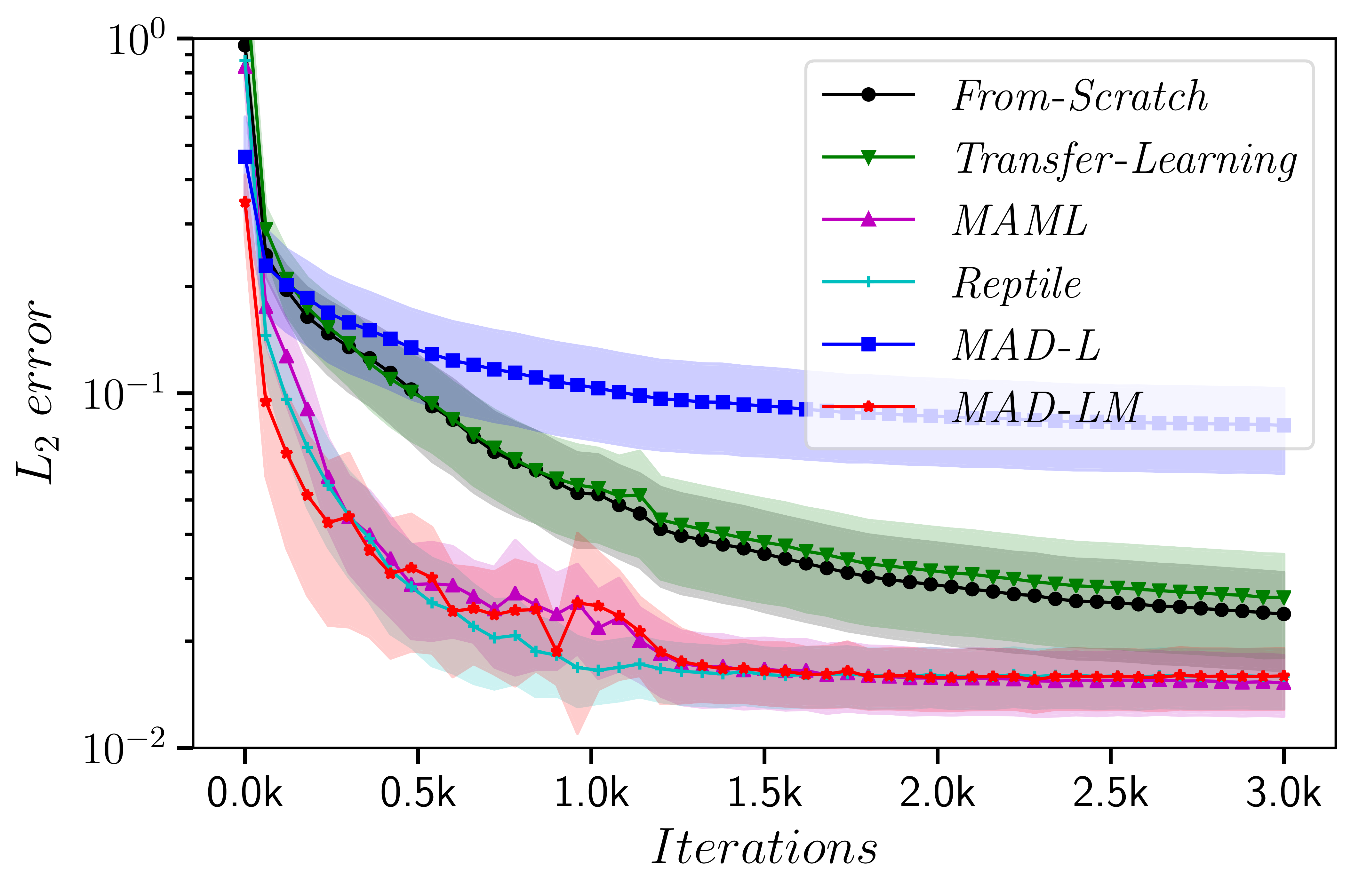}}
\caption{\textbf{Burgers' equation:} The mean $L_2\ error$ convergence with respect to the number of training iterations under heterogeneous PDE parameters.}
\label{fig:burgers_heterogeneous}
\end{center}
\end{figure}
For Burgers' equation, we also consider the scenario when the viscosity coefficients $\nu$ in Eq.\eqref{def:burgers} vary within a certain range, i.e., the PDE parameter $\eta=(\nu, u_0(x))$ is heterogeneous. 
Specifically, $\nu \in \{ 10^\beta | \beta \sim U(-3,-1) \}$ where $U$ is a uniform distribution and 
$u_0(x) \sim \mathcal{N} (0; 100(- \Delta + 9I)^{-3})$.
In this experiment, $|S_1|=100$ and $|S_2|=50$ while $S_1$ and $S_2$ come from the same task distribution.
Fig.\ref{fig:burgers_heterogeneous} compares the convergence curves of mean $L_2\ error$ corresponding to different methods.
\textit{MAD-LM} has obvious speed and accuracy improvement over \textit{From-Scratch} and \textit{Transfer-Learning}.
It's worth noting that \textit{MAML} and \textit{Reptile} also perform well in this scenario.

\begin{figure}
\centering
\subfigure[\textit{MAD-L}]{
	\label{fig:burgers_ablation_latent_size_MAD_l}
	\includegraphics[width=0.45\columnwidth]{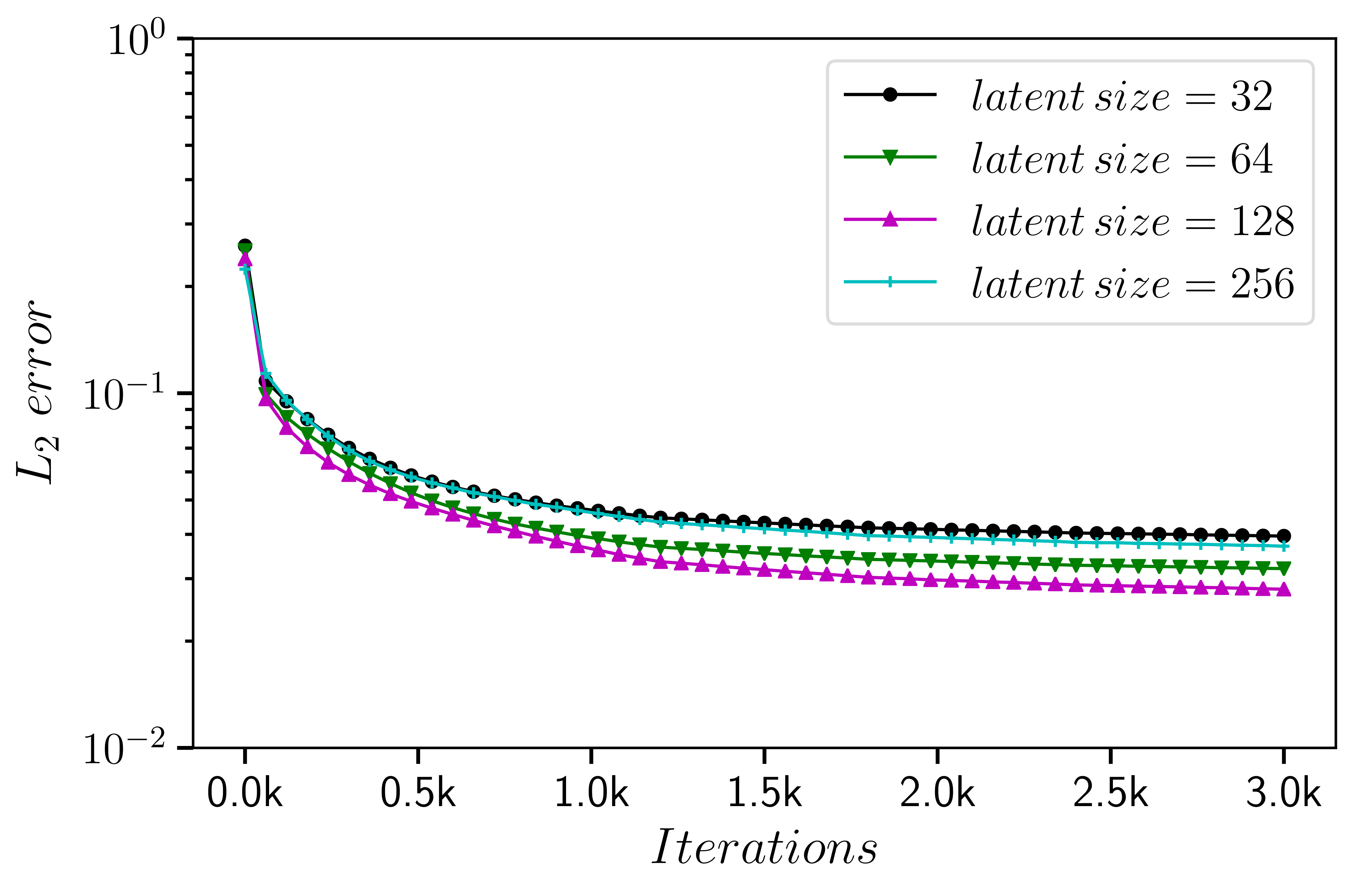}
}
\subfigure[\textit{MAD-LM}]{
	\label{fig:burgers_ablation_latent_size_MAD_lm}
	\includegraphics[width=0.45\columnwidth]{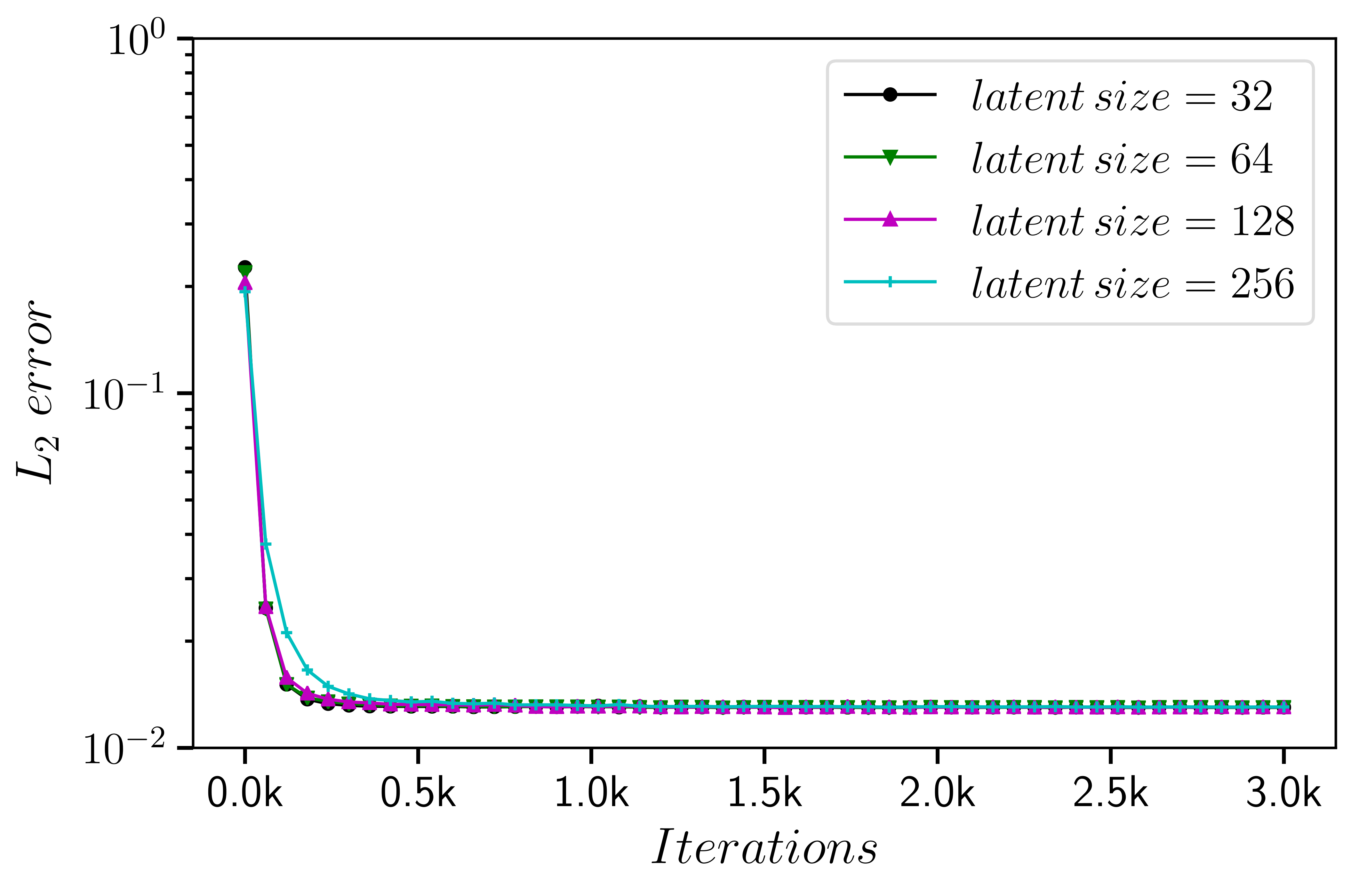}
}
\caption{\textbf{Burgers' equation:} The mean $L_2\ error$ of \textit{MAD-L} (a) and \textit{MAD-LM} (b) convergence with respect to the number of training iterations under different latent size.}
\end{figure}
We investigated the effect of the dimension of the latent vector (latent size) in Burgers' equation on performance. 
As can be seen from Fig.\ref{fig:burgers_ablation_latent_size_MAD_l}, for MAD-L, different latent sizes have different performances and the best performance is achieved when it is equal to 128. 
As can be seen from Fig.\ref{fig:burgers_ablation_latent_size_MAD_lm}, for MAD-LM, although these latent sizes are quite different, they achieve very close performance.

\section{Detailed Experimental Setup and Extended Results for Time-Domain Maxwell's Equations}\label{appendix:maxwell}
Except for the difference in equation coefficients $\epsilon_r$ and $\mu_r$, the settings of solution domain $\Omega$, initial conditions, boundary conditions and point source term $J$ are the same as those in \cite{huang2021solving}.
Specifically, the solution domain $\Omega$ is $[0, 1]^2 \times [0, 4e-9]$. 
The initial electromagnetic field is zero everywhere and the boundary condition is the standard Mur's second-order absorbing boundary condition \cite{schneider2010understanding}.
$J$ in Eq.\ref{eq:point_src} represents a known source function and we set it to a Gaussian pulse. 
In temporal, this function can be expressed as:
\begin{equation}
	\begin{aligned}
		J(x, y, t) = e^{-(\frac{t-d}{\tau })^2} \delta(x-x_0)\delta(y-y_0). \label{eq:gauss_pulse}
	\end{aligned}
\end{equation}
where $d$ is the temporal delay, $\tau$ is a pulse-width parameter, $\delta(\cdot)$ is the Dirac function used to represent the point source, and $(x_0, y_0)=(0.5,0.5)$ is the location of the point source, $\tau = 3.65 \times \sqrt{2.3}/(\pi f)$ and the characteristic frequency $f$ is set to be $1 GHz$.
To solve the singularity problem caused by the point source, we use the $\delta(\cdot)$ function approximation method and the lower bound uncertainty weighting method proposed by \cite{huang2021solving}. 
In addition, the MS-SIREN network structure proposed by \cite{huang2021solving} is used.
To measure the accuracy of the model, the reference solution is obtained through the finite-difference time-domain (FDTD) \cite{schneider2010understanding} method. 

We collect 25 pairs of $(\epsilon_r$, $\mu_r) $ at equal intervals from the region of $[1,5]^2 $ and randomly select 20 samples as $S_1$ with the rest 5 samples as $S_2$.
For the training of \textit{From-Scratch}, the pre-training and fine-tuning of \textit{Transfer-Learning}, we set the total number of iterations to $100$k.
For the pre-training and fine-tuning of \textit{MAD-L} and \textit{MAD-LM}, we set the total iterations to $200$k and $100$k, respectively.

\begin{figure}
\centering
\subfigure[\textit{MAD-L}]{
	\label{fig:maxwell_MAD_l}
	\includegraphics[width=0.48\columnwidth]{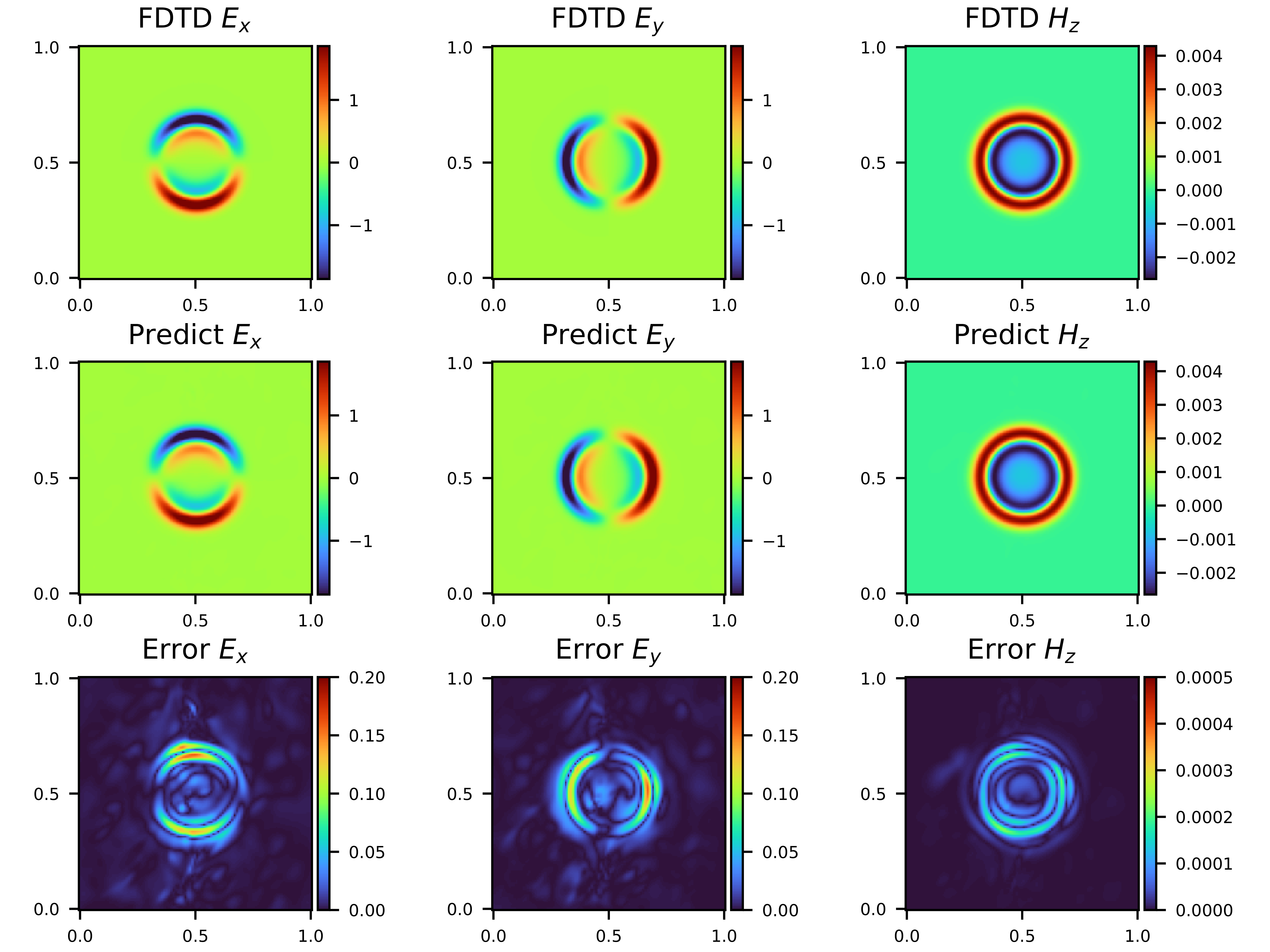}
}
\subfigure[\textit{MAD-LM}]{
	\label{fig:maxwell_MAD_lm}
	\includegraphics[width=0.48\columnwidth]{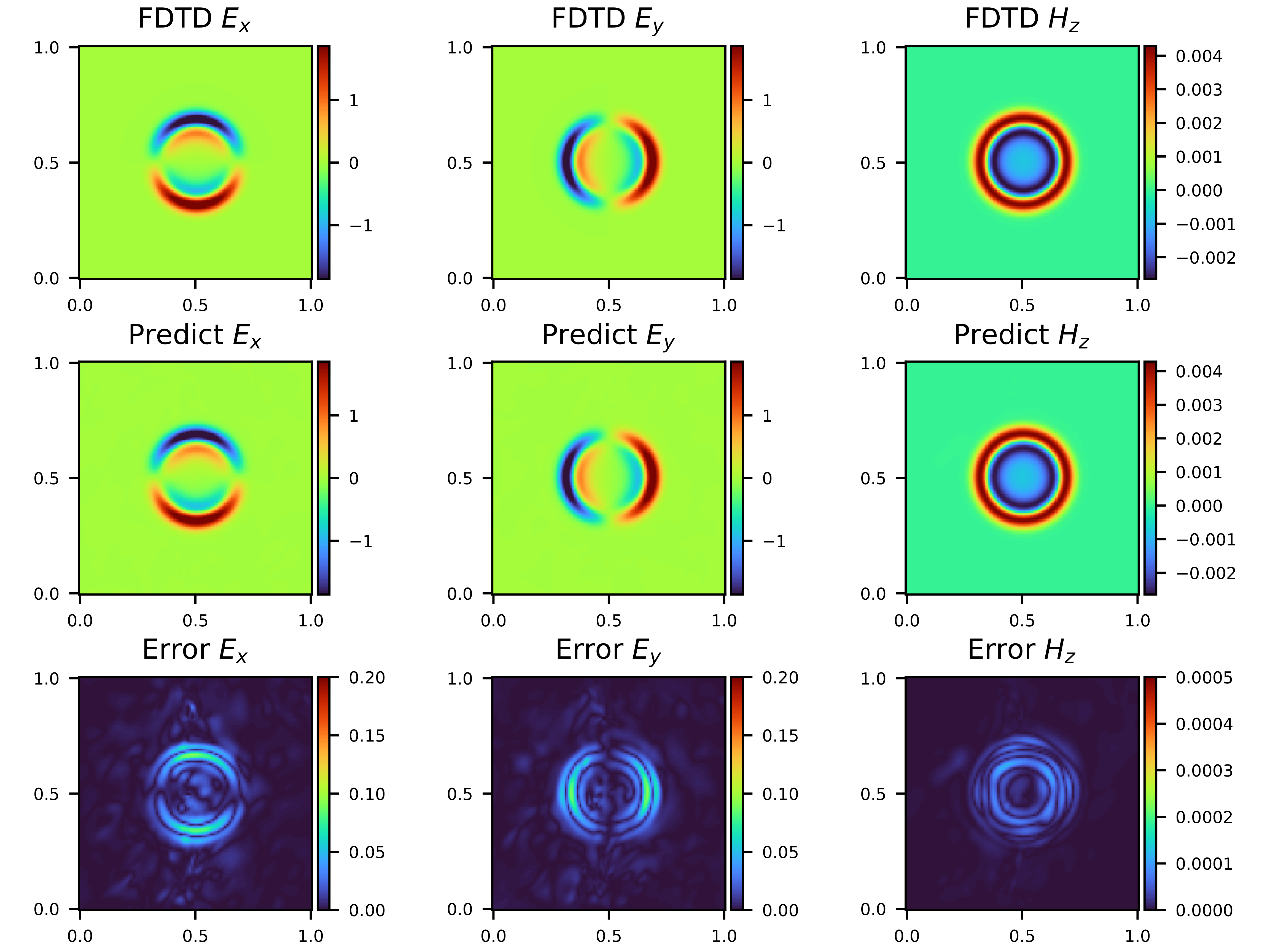}
}
\caption{
	\textbf{Maxwell's equations:} Model predictions of \textit{MAD-L} (a) and \textit{MAD-LM} (b) vs. the FDTD solutions at $t = 4ns$.
	\textbf{Top:} The numerical results of ($E_x, E_y, H_z$) by FDTD method;
	\textbf{Middle:} The predicted ($E_x, E_y, H_z$) through the deep learning model;
	\textbf{Bottom:} The absolute error between model predictions and the reference solutions.}
\label{fig:maxwell_result}
\end{figure}
The instantaneous electromagnetic fields at time $4ns$ of \textit{MAD-L} and \textit{MAD-LM} compared with the reference FDTD results when $(\epsilon_r$, $\mu_r) = (3, 5)$ are depicted in Fig.\ref{fig:maxwell_MAD_l} and Fig.\ref{fig:maxwell_MAD_lm}, respectively. 
By observing the absolute error between the model predictions and the reference FDTD results, we can see that \textit{MAD-LM} can achieve a lower absolute error. 
Specifically, the $L_2\ error$ of \textit{MAD-L} is 0.037 and that of \textit{MAD-LM} is 0.030.

We also apply \textit{PI-DeepONet} to the solution of time-domain Maxwell's equations with a point source. 
In this experiment, the branch net of \textit{PI-DeepONet} is a 4-layer fully connected network with 64 neurons in each hidden layer. 
The trunk net is an MS-SIREN~\cite{huang2021solving} network, which consists of 4 subnets, each with 7 fully connected layers and 64 neurons in each hidden layer.
We take the PDE parameter $(\epsilon_r,\mu_r)$ directly as the input of branch net. 
Because there are 3 output functions $\left( E_x(x,y,t), E_y(x,y,t), H_z(x,y,t) \right)$, we adopt the method proposed in \cite{LuLu2021ACA} to solve the multi-output problem, i.e. split the outputs of both the branch net and the trunk net into 3 groups, and then the $k$-th group outputs the $k$-th solution.
However, due to the optimization difficulties caused by the singularity of the point source, \textit{PI-DeepONet} appears to be very poor in accuracy (mean $L_2\ error$ is 0.672), while the mean $L_2\ error$ of \textit{MAD-LM} is 0.028.

\section{Detailed Experimental Setup and Extended Results for Laplace's Equation}\label{appendix:laplace}

We generated 150 samples using GRF and randomly selected 100 samples as $S_1$, the remaining 50 cases are used as $S_2$.
It should be declared that each sample corresponds to a different $h$ inside the boundary circle.
For each sample, we randomly select $16\times1024$ points from the interior of the triangle and obtain the analytic solutions corresponding to these points to evaluate the accuracy of the model.
When we apply the MAD method to solve this problem, it is not convenient to measure the distance between $\eta_{\text{new}}$ and $\eta_i$ since the variable PDE parameter $\eta$ is the shape and boundary condition of the solution domain.
Therefore, the average of $|S_1|$ latent vectors obtained in the pre-training stage is used as the initialization of $z$ in the fine-tuning stage.
For the training of \textit{From-Scratch}, the pre-training and fine-tuning of \textit{Transfer-Learning}, we set the total number of iterations to $10$k. 
For the pre-training and fine-tuning of \textit{MAD-L} and \textit{MAD-LM}, we set the total iterations to $50$k and $10$k, respectively.

\begin{figure}
\begin{center}
\centerline{\includegraphics[width=0.8\columnwidth]{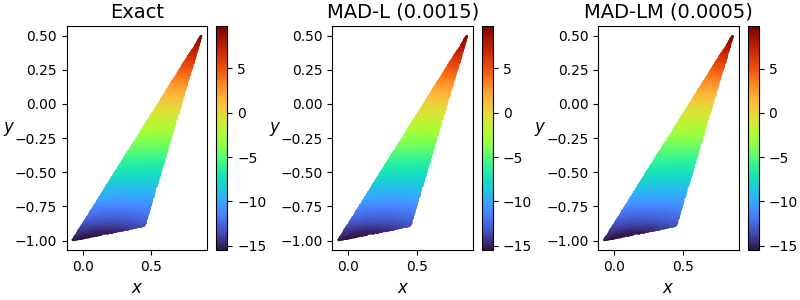}}
\caption{\textbf{Laplace's equation:} Analytical solutions, model predictions of \textit{MAD-L} and \textit{MAD-LM} (left to right).}
\label{fig:laplace_result}
\end{center}
\end{figure}

Fig.\ref{fig:laplace_result} shows the predictions of \textit{MAD-L} and \textit{MAD-LM} compared with the analytical solutions under a randomly selected sample in $S_2$. 
To the naked eye, the prediction results of \textit{MAD-L} and \textit{MAD-LM} are almost identical to those of the analytical solution.
However, the $L_2\ error$ of \textit{MAD-L} is 0.0015, and that of \textit{MAD-LM} is 0.0005.


\begin{figure}
\begin{center}
	\subfigure[\textit{Entire Fine-tuning Process}]{
		\label{fig:laplace_polygon_a}
		\includegraphics[width=0.48\columnwidth]{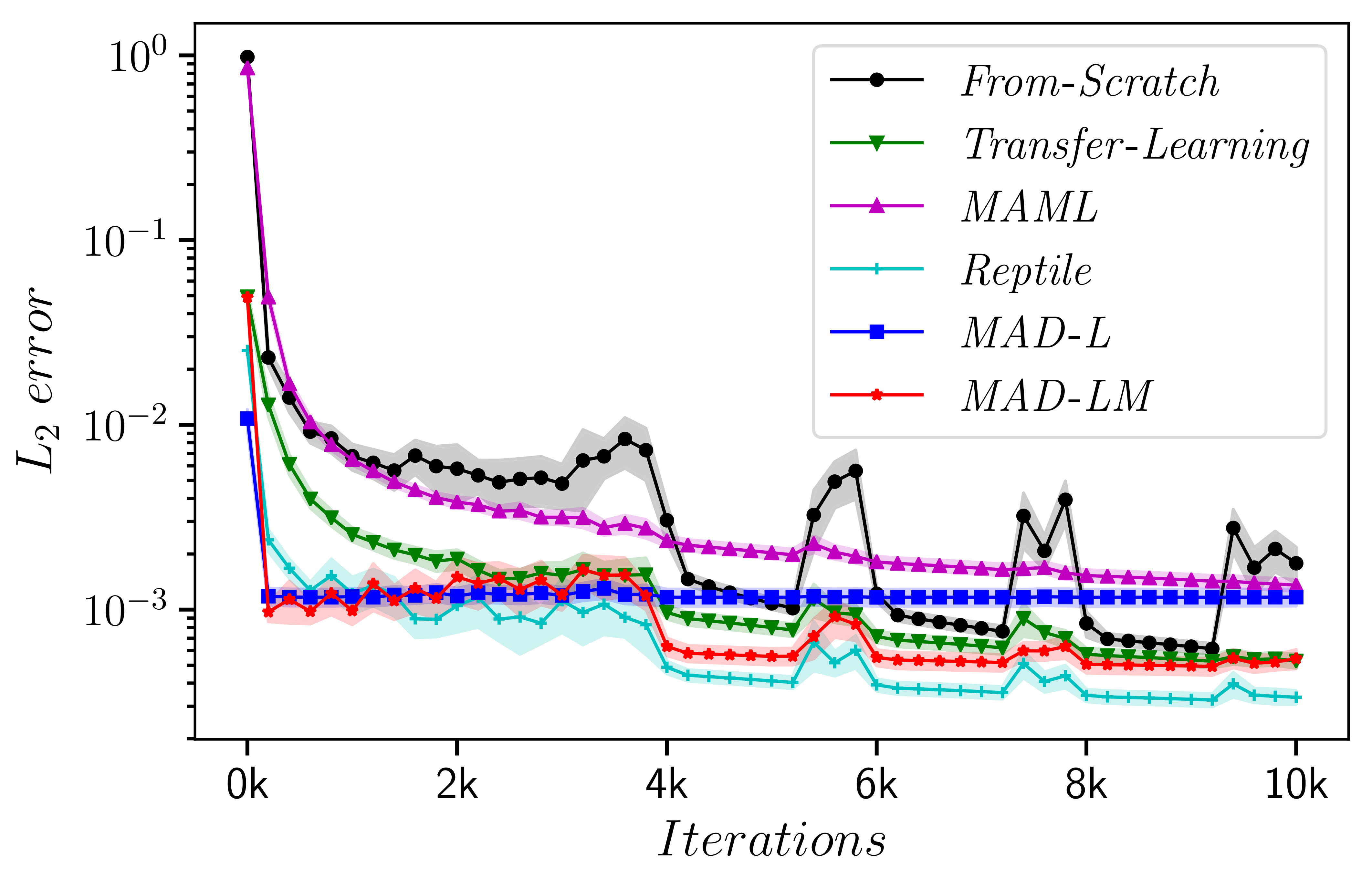}
	}
	\subfigure[\textit{First 1000 Iterations}]{
		\label{fig:laplace_polygon_b}
		\includegraphics[width=0.48\columnwidth]{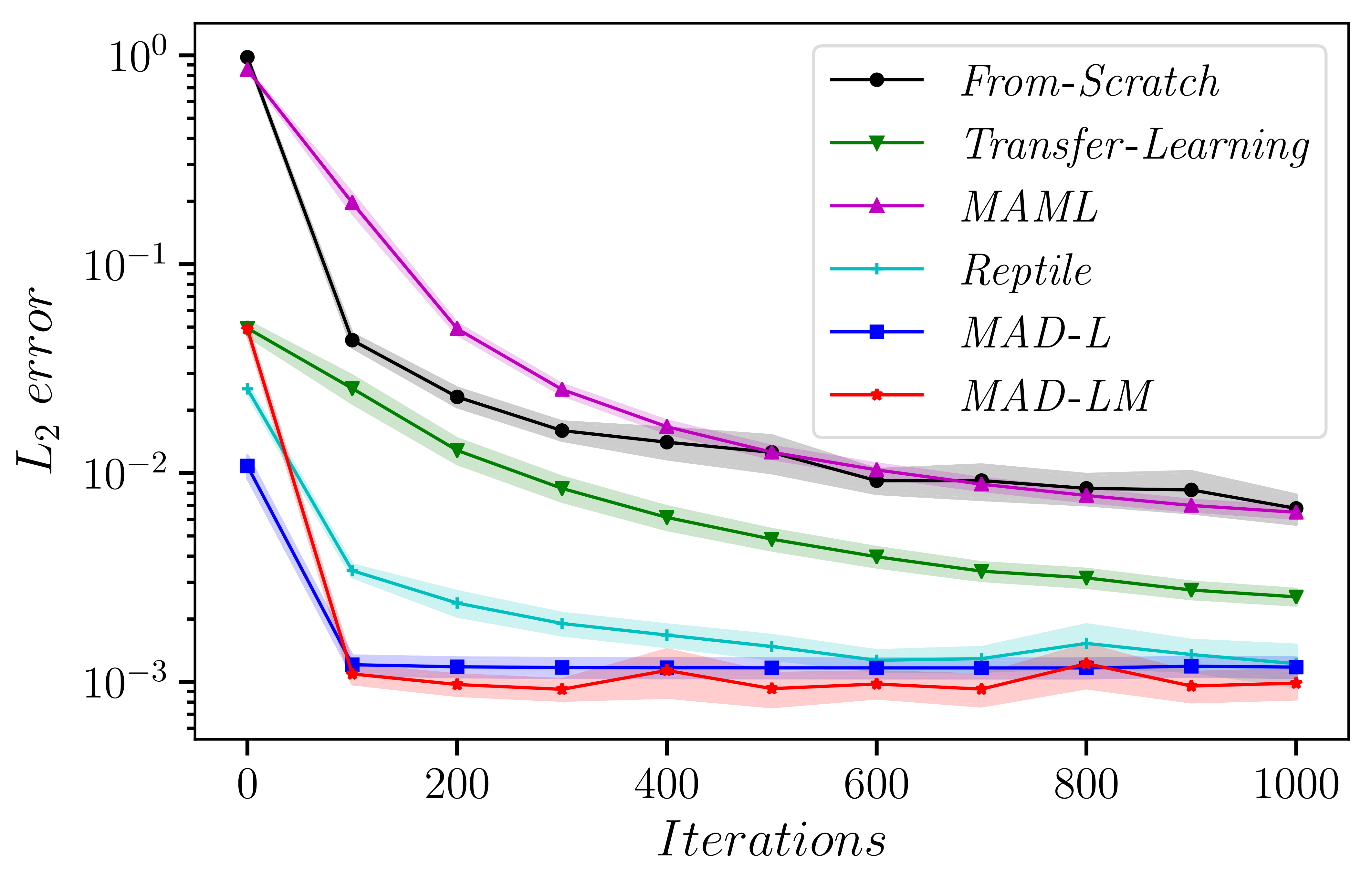}
	}
\caption{\textbf{Laplace's equation:} The mean $L_2\ error$ convergence with respect to the number of training iterations when the solution domain $\Omega$ is a polygon with different shapes.}
\label{fig:laplace_polygon}
\end{center}
\end{figure}

\begin{figure}
\begin{center}
\centerline{\includegraphics[width=0.8\columnwidth]{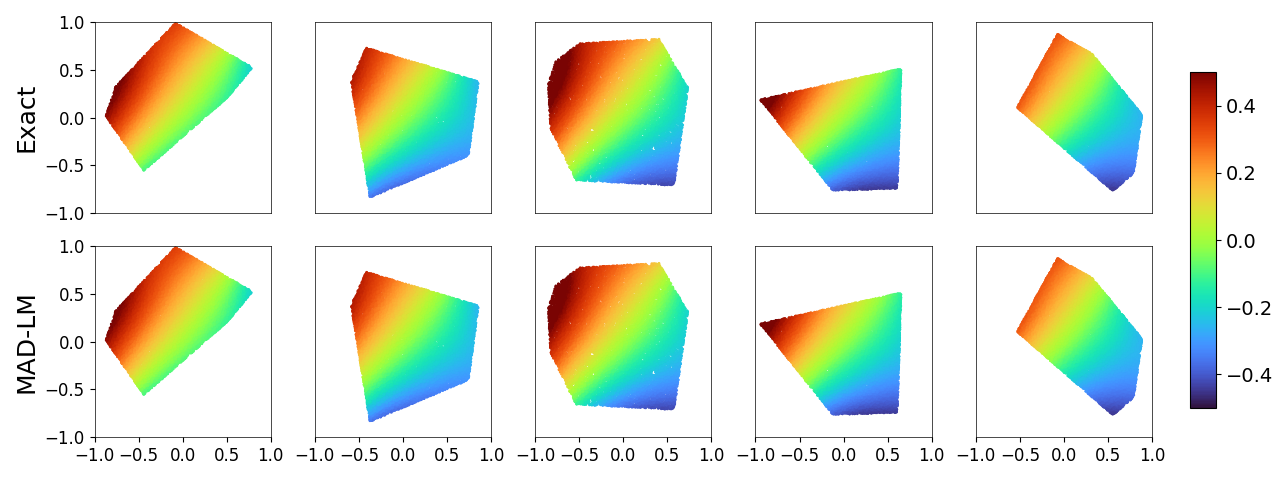}}
\caption{\textbf{Laplace's equation:} Analytical solutions, model predictions of \textit{MAD-LM} when the solution domain $\Omega$ is a polygon with different shapes.}
\label{fig:laplace_polygon_result}
\end{center}
\end{figure}

We also consider a more complex scenario for Laplace's equation. 
Specifically, the shape of the solution domain $\Omega$ is a convex polygon arbitrarily taken from the interior of the unit circle. 
The number of sides of the polygon is in the range [3, 10] and the boundary conditions of the polygon are generated in the same way as in Sec.\ref{sec:laplace}.
It should be emphasized that the PDE parameters are heterogeneous for all experiments of Laplace's equation.
Therefore, different solution domain shapes correspond to different $h$ and different $g(x,y)$.
In this experiment, $|S_1|=100$ and $|S_2|=50$.
Fig.\ref{fig:laplace_polygon} compares the convergence curves of mean $L_2\ error$ corresponding to different methods, 
and Fig.\ref{fig:laplace_polygon_a} shows the entire fine-tuning process and Fig.\ref{fig:laplace_polygon_b} zooms in on the results of the first 1000 iterations. 
Compared to other methods, \textit{MAD-L} and \textit{MAD-LM} can achieve faster adaptation, i.e. very low $L_2\ error$ in less than 100 iterations.
Fig.\ref{fig:laplace_polygon_result} shows a comparison of the prediction of \textit{MAD-LM} with the analytical solution under 5 randomly selected samples in $S_2$.

\begin{figure}
\begin{center}
	\subfigure[\textit{Entire Fine-tuning Process}]{
		\label{fig:laplace_ellipse_a}
		\includegraphics[width=0.48\columnwidth]{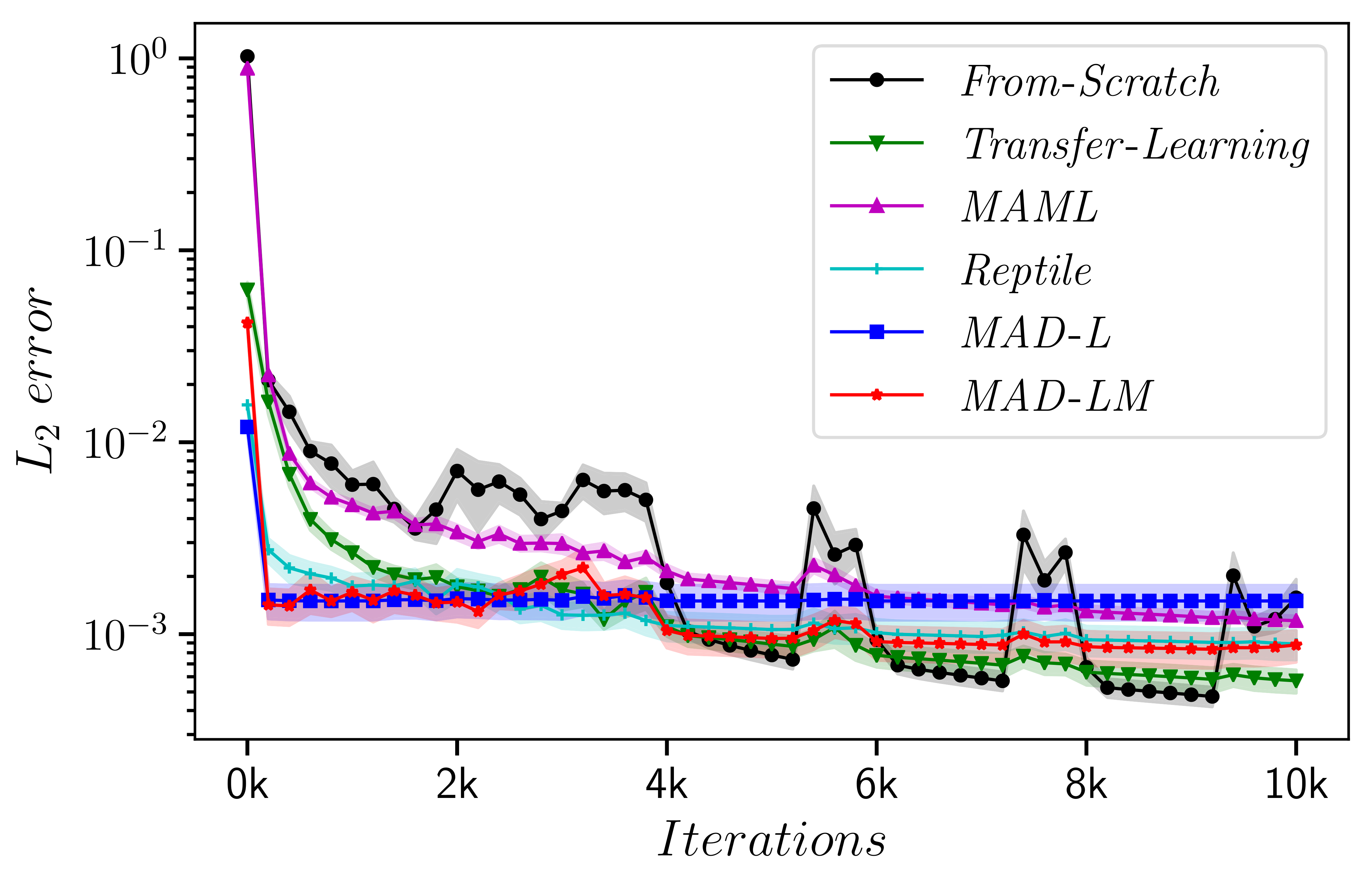}
	}
	\subfigure[\textit{First 1000 Iterations}]{
		\label{fig:laplace_ellipse_b}
		\includegraphics[width=0.48\columnwidth]{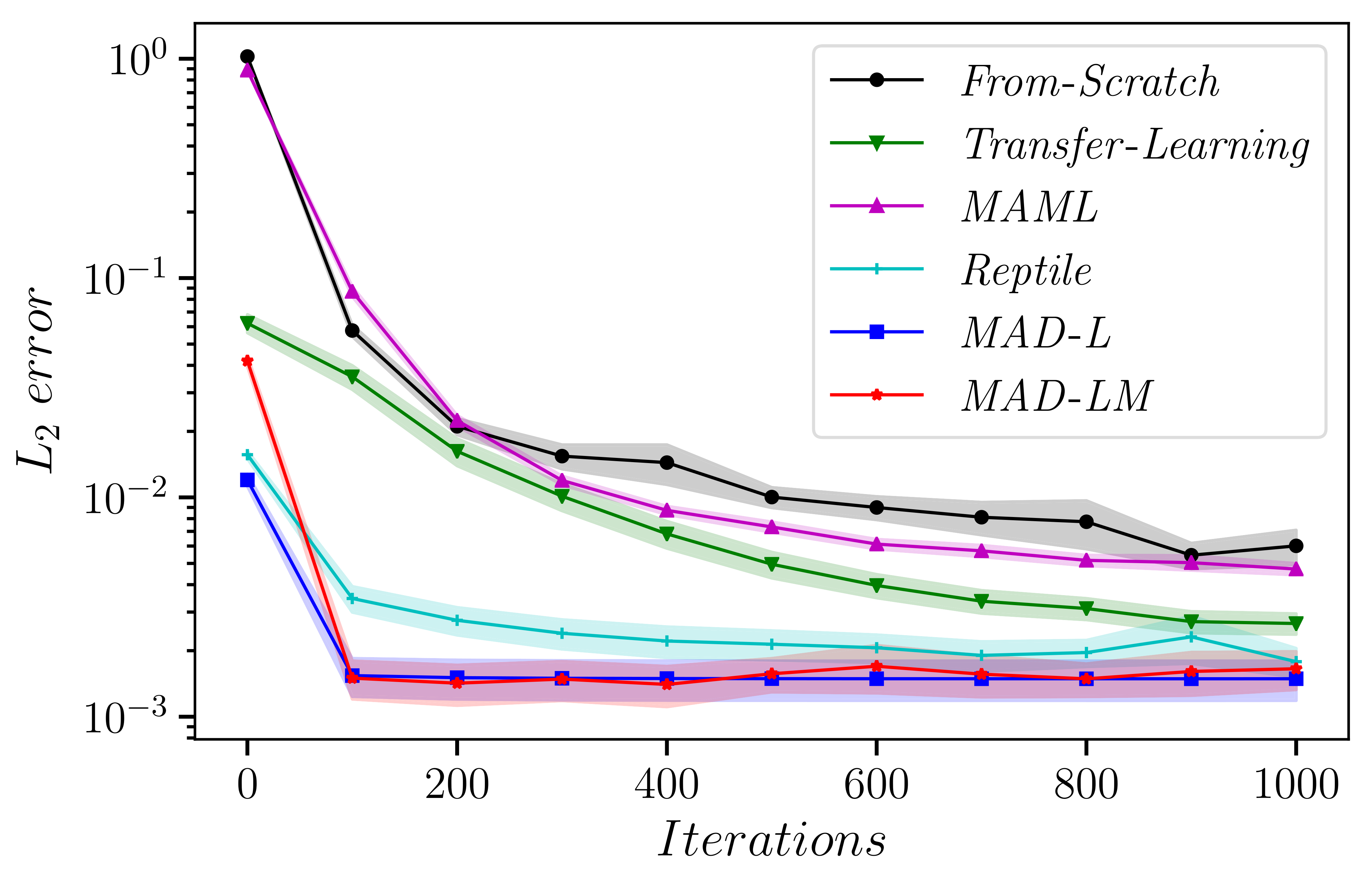}
	}
\caption{\textbf{Laplace's equation:} The mean $L_2\ error$ convergence with respect to the number of training iterations for extrapolation experiments.}
\label{fig:laplace_ellipse}
\end{center}
\end{figure}

\begin{figure}
\begin{center}
\centerline{\includegraphics[width=0.8\columnwidth]{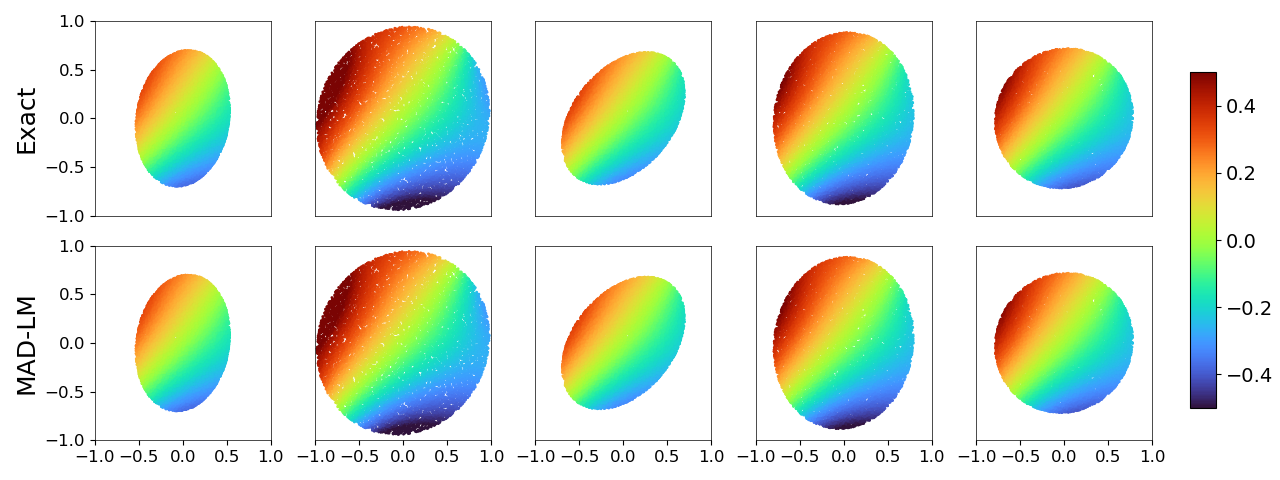}}
\caption{\textbf{Laplace's equation:} Analytical solutions, model predictions of \textit{MAD-LM} for extrapolation experiments.}
\label{fig:laplace_ellipse_result}
\end{center}
\end{figure}

We also do an extrapolation experiment for Laplace's Equation.
Specifically, in the pre-training stage, the shape of the solution domain $\Omega$ in $S_1$ is a convex polygon arbitrarily taken from the interior of the unit circle. 
However, in the fine-tuning stage, the shape of the solution domain $\Omega$ in $S_2$ is an ellipse arbitrarily taken from the interior of the unit circle.
In this experiment, $|S_1|=100$ and $|S_2|=20$.
Fig.\ref{fig:laplace_ellipse_b} shows that even in the case of extrapolation, \textit{MAD-LM} can show faster adaptation compared to other methods.
Fig.\ref{fig:laplace_ellipse_result} shows a comparison of the prediction of \textit{MAD-LM} with the analytical solution under 5 randomly selected samples in $S_2$, which demonstrates the high accuracy of the solution obtained by \textit{MAD-LM}.

\end{document}